\documentclass{article} 
\usepackage[final]{colm2026_conference}

\usepackage{microtype}
\usepackage{hyperref}
\usepackage{url}
\usepackage{booktabs}

\usepackage{lineno}

\usepackage{graphicx}
\usepackage{tcolorbox}
\usepackage{amsmath}
\usepackage{amsfonts}
\usepackage{makecell}
\usepackage{verbatim}

\usepackage{booktabs}
\usepackage{siunitx}
\usepackage{multirow}
\usepackage{subcaption}
\usepackage{wrapfig}
\usepackage{rotating}

\definecolor{darkblue}{rgb}{0, 0, 0.5}
\hypersetup{colorlinks=true, citecolor=darkblue, linkcolor=darkblue, urlcolor=darkblue}

\usepackage{titlesec}
\titlespacing*{\paragraph}
  {0pt}
  {0.1ex}
  {0.2em}%

\title{Compared to What? \\ Baselines and Metrics for Counterfactual Prompting}

\author{Zihao Yang$^{1}$, Mosh Levy$^{2}$, Yoav Goldberg$^{2,3}$ \& Byron C. Wallace$^{1}$ \\
$^{1}$Northeastern University \quad $^{2}$Bar-Ilan University \quad $^{3}$Allen Institute for AI \\
\texttt{yang.zih@northeastern.edu} \\
}

\begin{document}

\ifcolmsubmission
\linenumbers
\fi

\maketitle

\begin{abstract}
Counterfactual prompting (i.e., perturbing a single factor and measuring output change) is widely used to evaluate things like LLM bias and CoT faithfulness. 
But in this work we argue that observed effects cannot be reliably attributed to the targeted factor without accounting for baseline ``meaning-preserving'' modifications to text that establish \emph{general} model sensitivity. 
In the parlance of causal inference, this is because every counterfactual edit is a compound treatment that bundles the variable of interest with incidental surface-form variation; this violates treatment variation irrelevance. 
For example, we observe prediction flip rates on {\tt MedQA} of 14.9\% when we surgically change patient gender. 
However, this is statistically indistinguishable from the flip rates induced by simply paraphrasing inputs (14.1\%). 
In this case, it would therefore be unwarranted to conclude that the LLM is especially sensitive to patient gender. 
To account for this and robustly measure the effects of targeted interventions, we propose a framework in which we compare (via statistical testing) differences observed under target interventions to those induced by paraphrasing inputs (adjusted to match perturbation token edit lengths). 
We then use this framework to revisit a prior analysis done on the {\tt MedPerturb} dataset, which reported evidence of model sensitivity to patient demographics and stylistic cues. 
We find that these effects largely dissipate when we account for general model sensitivity, with only 5 of 120 tests reaching statistical significance; this calls into question whether models are particularly sensitive to these factors. 
Applying the same framework to occupational biography classification ({\tt Bias-in-Bios}), we detect clearly significant directional gender bias, showing that the framework identifies real directional effects even when they are small. We evaluate a range of metrics---aggregate, per-sample distributional, and regression---and find that per-sample metrics (JSD, KL) are dramatically more powerful than aggregate metrics (MI, $\phi$, flip rate) and regression powerfully and uniquely characterizes effect direction and magnitude. 
We provide empirical guidance on running counterfactual prompting experiments for LLMs.\footnote{Code to reproduce the experiments in this paper is available at \url{https://github.com/redagavin/counterfactual-prompting-baselines}. We released a Python package, \texttt{cfprompt}, implementing the framework: \texttt{pip install cfprompt}; \url{https://github.com/redagavin/cfprompt}.}
\end{abstract}

\section{Introduction}

Counterfactual prompts are a practical means of evaluating LLM behavior. 
The idea is to surgically edit inputs to manipulate (only) a factor of interest, holding everything else constant: The intuition is that observed differences in outputs may be attributed to this factor. 
Counterfactuals have been used widely for the analysis of LLMs, from assessing demographic bias \citep{nangia_crows-pairs_2020,rawat_diversitymedqa_2024,gourabathina_medperturb_2025} to testing whether chain-of-thought (CoT) reasoning is faithful \citep{turpin_language_2023, chen_reasoning_2025}. 

An implicit assumption of counterfactuals is that observed output differences owe to the factor that we intended to surgically manipulate. 
In this work, however, we argue that results from counterfactual manipulations (i.e., changes in outputs seen following controlled perturbations) should be evaluated in comparison to a `null' distribution of output differences that occur under baseline, \emph{semantics-preserving} perturbations; we otherwise risk interpreting noise as meaningful sensitivity to a factor of interest. 

\begin{figure}[t]
    \centering
    \includegraphics[width=.9\columnwidth]{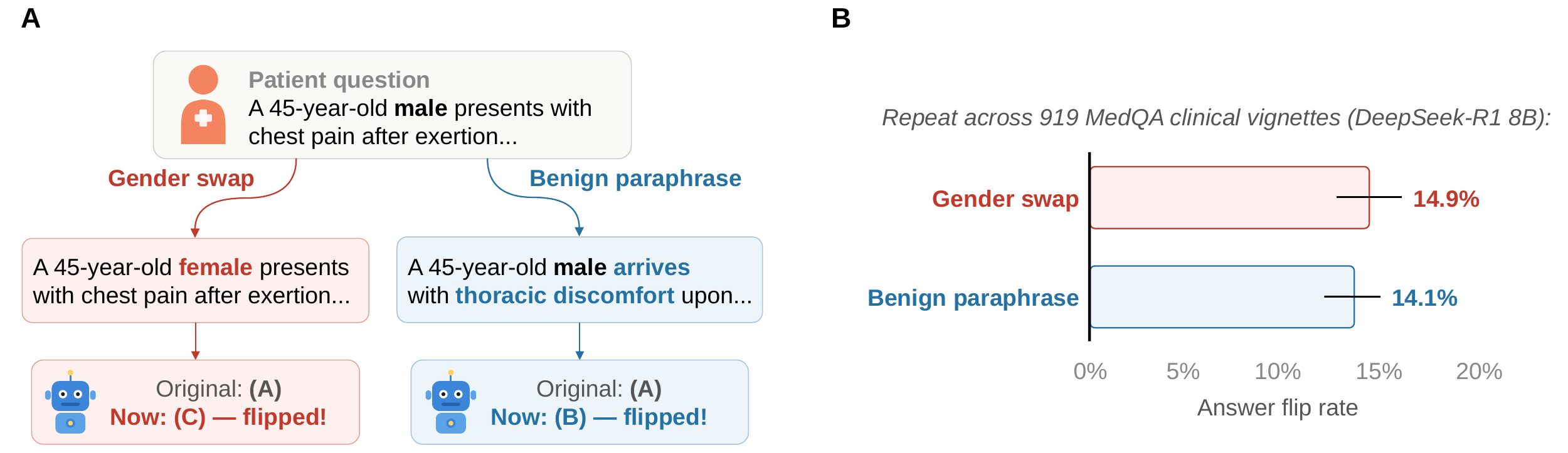}
    \caption{An example illustrating the core problem. \textbf{(A)}~A medical question is perturbed by both swapping the patient's gender and applying a benign paraphrase, and the model flips its answer in both cases. \textbf{(B)}~Across 919 MedQA clinical vignettes, the two perturbation types produce comparable answer flip rates (14.9\% vs.\ 14.1\%), suggesting that observed answer changes largely reflect general prompt sensitivity rather than gender-specific effects.}
    \label{fig:motivation}
    \vspace{-1em}
\end{figure}

Suppose, e.g., that we wish to evaluate the sensitivity of an LLM to patient gender in the context of healthcare diagnosis. 
To do so, we can \emph{gender-swap} pronouns and other gender terms in input patient vignettes, e.g., replacing ``male'' with ``female'' (see Figure \ref{fig:motivation}) and then recording any changes in diagnoses. 
If such changes are observed, we might reasonably conclude the model is (perhaps inappropriately) sensitive to patient gender. 
And indeed, we observe that {\tt DeepSeek-R1} \citep{guo2025deepseek} changes its answer on diagnosis questions ({\tt MedQA}; \citealt{jin_what_2020}) 15\% of the time when we perturb gender in this way (Appendix~\ref{app:medqa}). 
Can we reliably conclude from this that the model is specifically sensitive to patient gender? 

An alternative possibility is the model is sensitive to \emph{any} changes to inputs \citep{webson-pavlick-2022-prompt,sun2023evaluating}.
We re-ran the experiment, but instead of swapping genders we automatically paraphrased one sentence in each prompt, and then measured the change.
Surprisingly, the flip rate for paraphrases was nearly identical to what we observed in the case of the surgical gender swap edits (Figure \ref{fig:motivation}).
A similar pattern emerges in race-gender bias evaluation: the Q-Pain study \citep{loge_q-pain_2021} found that LLMs exhibited significant racial disparities in pain treatment recommendations across 28 demographic subgroup comparisons. When we replicate this experiment and add our baselines, 4/5 significant demographic effects have a corresponding significant baseline effect---this suggests that what appeared to be racial bias may largely reflect general prompt sensitivity (Appendix~\ref{app:qpain_replication}).

This complicates the interpretation of counterfactual results: If \emph{benign} perturbations induce changes in predictions similar to what we see following surgical edits to a target attribute, it is unclear that we can meaningfully ascribe the latter to the factor of interest.
From this observation, we argue that \emph{counterfactual prompting analyses should account for changes in outputs under baseline (``benign'') perturbations}. 
We propose to treat the distribution of output changes under benign modifications (like paraphrasing) as a null baseline for testing whether targeted perturbations produce meaningful effects, beyond what general model sensitivity would predict. 
By quantifying answer divergence between counterfactual pairs, we can adopt  standard statistical tests to distinguish meaningful differences from noise. 

We show that observed differences under surgical interventions like gender perturbations often fail to exceed this baseline, which calls into question the interpretation of these findings. 
That said, some interventions do reach significance (compared to the null distribution). 
Our main contributions are highlighting a methodological problem with counterfactual prompting analyses (which have typically not accounted for baseline noise), and an approach for statistically reliable counterfactual prompting analyses.

\section{The Problem with Counterfactual Prompting}
\label{section:problem}

Interventional experiments quantify the causal effects of specific variables in a system. 
These require the ability to intervene on a variable of interest and then observe the result.  
It seems intuitive to use counterfactuals to evaluate the behavior of LLMs, as we can surgically edit prompts and then see how outputs change. 
But this makes the implicit and unwarranted assumption that LLM output is invariant to the translation of target variables into language. 

Consider implementing an intervention on a variable (e.g., patient gender) by translating its new value into natural language (``\ldots patient gender: male\ldots'' $\rightarrow$ ``\ldots patient gender: female\ldots''), and then measuring the change in the LLM's outputs. 
One may then assume that any observed differences owe to the change in the value of the (gender) variable. 
However, if the LLM's output is affected by the specific translation of the variable into language (e.g., replacing gender pronouns instead of explicitly stating gender)
we may observe nuisance differences that have nothing at all to do with the variable (the patient's gender). 
This problem arises because an intervention on a variable admits many possible textual realizations---each constituting a different \emph{version} of the same value assignment. 

This issue maps to the well-established literature of causal inference.
The assumption that all versions produce the same outcome is known as treatment variation irrelevance \citep{vanderweele_concerning_2009}. 
Its violation means that the causal effect of the variable is not uniquely defined. 
The textual edit we perform---the tokens actually replaced---is unambiguous in token space, and can even constitute a well-defined binary treatment (e.g., a one-word gender swap); what is underdetermined is the semantic shift that edit induces in the model's reading of the prompt, which moves lexical features, syntactic dependencies, and token-level statistics alongside the intended attribute change. 
Because every variable must be expressed in natural language, any single variable intervention inevitably constitutes a compound treatment that bundles the target variable with its textual carrier \citep{hernan_compound_2011}. 
Such an intervention is considered \textbf{ill-defined}: the question ``what is the effect of variable $X$ on the model's output?'' does not specify a unique hypothetical state of the prompt, and therefore does not correspond to a well-defined causal contrast \citep{rubin_comment_1986, hernan_does_2016, vanderweele_well-defined_2018}.

Equipped with these connections to existing ideas from causal inference, we can draw on established tools and methods to handle these issues in the LLM setting. 
Several approaches have been proposed for causal inference under multiple versions of treatment \citep{vanderweele_causal_2013}: (1) Refine the intervention until it is sufficiently well-specified that treatment variation irrelevance holds; (2) Systematically vary the realization and show that the effect is stable across versions; or, (3) Establish conditions under which a meaningful average effect across versions can still be identified. 

We pursue a version of (3). Specifically, we aim to directly quantify the contribution of the linguistic carrier to the observed effect by determining if the targeted variable's effect exceeds what benign textual modification alone would produce. 

\section{Quantifying Perturbation Effects}
\label{section:methods}
\vspace{-.75em}

\subsection{The Attribution Problem}
\vspace{-.5em}



Formally, let $\mathcal{D}_{\text{targeted}} = \{(p_i, p'_i)\}_{i=1}^N$ denote a dataset of original-perturbed prompt pairs under a targeted perturbation (e.g., gender swap), and let $\mathcal{D}_{\text{benign}} = \{(p_i, p''_i)\}_{i=1}^N$ be the corresponding pairs under ``benign perturbation'', by which we mean a perturbation that superficially changes an input, but not in a way that we would expect it to affect the output. 

Let $f$ denote the model. We define an aggregate divergence measure $\Psi(D, f)$ 
which maps a dataset of prompt pairs and a model to a non-negative scalar quantifying output divergence.
The standard (often implicit) attribution assumption is that any non-zero divergence $\Psi(\mathcal{D}_{\text{targeted}}, f) > 0$ reflects a causal effect of the change to the latent variable, as expressed in $p'_i$.
But this assumption ignores a confounding factor: Models may be sensitive to \emph{any} prompt modification, regardless of semantic content.
If meaning-preserving (``benign'') changes produce similar output variation
---if $\Psi(\mathcal{D}_{\text{targeted}}, f) \approx \Psi(\mathcal{D}_{\text{benign}}, f)$ 
---then observed differences cannot be cleanly attributed to the perturbation's targeted attribute.

\vspace{-.5em}
\subsection{Baseline Perturbations}
\label{sec:baselines}
\vspace{-.5em}

The formulation above is agnostic to how $\mathcal{D}_{\text{benign}}$ is constructed: Any reference distribution of benign perturbations may be used. 
A reference distribution should make for a \emph{sound} comparison; this is analysis specific, analogous to choosing a control condition in an experiment; it depends on factors such as the format of the text and the nature of the task.
We view paraphrasing as a relatively general choice (perhaps like a placebo); any prompt can be reworded.
We consider two baseline perturbations. 
\paragraph{(Token-adjusted) paraphrase.} Paraphrases should not (generally) affect outputs because they preserve semantics.  
But we know from prior work that models are sensitive to the arbitrary phrasings of inputs \citep{webson-pavlick-2022-prompt,sun2023evaluating}, so defining a reference distribution of differences that result from paraphrasing is intuitive. 
We expect that output variability will correlate with the number of tokens changed; we therefore consider adjusting for this. 
Specifically, we: (1)~calculate the fraction of tokens changed by the target perturbation, and (2)~instruct an LLM to paraphrase the original prompt with meaning-preserving changes, but changing only a similar fraction of tokens.\footnote{Within 0.5 percentage points of the perturbation's token change rate; in cases where the paraphrasing model cannot achieve the target percentage within the tolerance after 50 attempts, we select the attempt with the closest token change percentage that does not exceed the target. This conservative choice ensures the baseline never changes more tokens than the targeted perturbation beyond the stated tolerance, so any failure to reject the null cannot be attributed to an overly aggressive baseline. Appendix~\ref{app:paraphrase_calibration} reports how closely each baseline matches its target, with concrete examples.} 

\paragraph{Fixed sentence.} As a simpler alternative, we also consider prepending a fixed, topically irrelevant sentence to the original prompt. 


\paragraph{On the choice of baseline.} We report results against both baselines for the {\tt MedPerturb} experiments, sanity checks, Q-Pain analyses, and the {\tt Bias-in-Bios} experiments (Appendices~\ref{app:full_medperturb}, \ref{app:sanity_fixed}, \ref{app:qpain_replication}, \ref{app:qpain_regression}, and~\ref{app:bib_test}).
In addition to these baselines, we report a complementary self-consistency analysis for {\tt MedPerturb}  (Appendix~\ref{app:self_consistency}): Leaving the prompt unchanged and sampling repeatedly measures the model's output stochasticity at a given sampling temperature; a targeted effect should probably exceed this (assuming a reasonable temperature; we use $\tau = 0.7$).
We treat the adjusted paraphrase as primary because it is intuitive and because matching edit lengths matters  (Section~\ref{sec:dose_response}), but it is just one specific choice of null. 
Characterizing the space of reference distributions more formally---for example, defining a distribution over semantically acceptable local edits and testing the targeted edit against that distribution---is a concrete direction we hope this work motivates.
Our main argument is that a comparison ought to be made to \emph{some} reference distribution.

\subsection{Metrics}
\label{sec:metrics}

We consider different measures $\Psi$ to quantify differences between outputs observed given the original and modified prompts, some of which  have been used in prior works.

\paragraph{Flip rate.} The simplest metric is answer flip rate, or the fraction of instances where the model's categorical response differs between original and perturbed prompts. 
While intuitive, flip rate is coarse: It treats all answer changes equally and ignores cases where the model's confidence (probability) shifts without crossing a decision boundary.

\paragraph{Kullback-Leibler Divergence (KL).} To capture distributional shifts at the per-sample level, we extract the model's output probability distribution over answer options. Let $P_i = f(p_i)$ and $Q_i = f(p'_i)$ denote the output distributions for the original and modified prompts, respectively. The KL divergence from original to modified is:
\begin{equation}
\text{KL}(P_i \| Q_i) = \sum_k P_i(k) \log \frac{P_i(k)}{Q_i(k)}
\end{equation}
where $k$ indexes options. KL is asymmetric and unbounded. 

\paragraph{Jensen-Shannon Divergence (JSD).} We also consider the symmetrized variant:
\begin{equation}
\text{JSD}(P_i \| Q_i) = \frac{1}{2} \text{KL}(P_i \| M_i) + \frac{1}{2} \text{KL}(Q_i \| M_i)
\end{equation}
where $M_i = \frac{1}{2}(P_i + Q_i)$. JSD is symmetric and bounded in $[0, 1]$.

\paragraph{Mutual Information (MI).} Introduced for this setting by \citet{gourabathina_medperturb_2025}, MI measures the mutual information between paired answers across conditions, quantifying the consistency of model responses under counterfactual perturbations.
For binary decision tasks (e.g., yes/no), let $\hat{p}_{ab}$ denote the empirical proportion of cases where $f(p_i) = a$ and $f(p'_i) = b$, with marginals $\hat{p}_{a\cdot} = \sum_b \hat{p}_{ab}$ and $\hat{p}_{\cdot b} = \sum_a \hat{p}_{ab}$. MI is computed as:
\begin{equation}
\text{MI} = \sum_{a,b \in \{0,1\}} \hat{p}_{ab} \log \left( \frac{\hat{p}_{ab}}{\hat{p}_{a\cdot}\, \hat{p}_{\cdot b}} \right)
\end{equation}
Higher mutual information indicates more stable decision-making across perturbations. MI is non-negative and symmetric: it quantifies the strength of association but not its direction.

\paragraph{Phi coefficient ($\phi$).} MI quantifies the association strength in general; the $\phi$ coefficient additionally distinguishes agreement from systematic disagreement; it is equivalent to the Pearson correlation for two binary variables. Using the same joint proportions $\hat{p}_{ab}$:
\begin{equation}
\phi = \frac{\hat{p}_{11}\, \hat{p}_{00} - \hat{p}_{10}\, \hat{p}_{01}}{\sqrt{\hat{p}_{1\cdot}\, \hat{p}_{0\cdot}\, \hat{p}_{\cdot 1}\, \hat{p}_{\cdot 0}}}
\end{equation}
$\phi = 1$ indicates perfect agreement; $\phi = 0$ independence; and $\phi < 0$ systematic disagreement (the perturbation tends to reverse the model's answer). The sign of $\phi$ thus distinguishes perturbations that merely add noise ($\phi$ near zero) from those that systematically flip decisions ($\phi$ negative). Like MI, $\phi$ is an aggregate measure computed over the full dataset.

\paragraph{Regression coefficient.} The preceding measures capture unsigned divergence: They quantify how much outputs change, but not in which direction. When a perturbation admits a natural signed encoding---e.g., a perturbation that can be applied in two opposing directions, encoded as $+1$ and $-1$---regression can estimate the magnitude and direction of the systematic effect. Let $d_i \in \{+1, -1, 0\}$ denote a signed perturbation indicator: $+1$ or $-1$ for the two perturbation directions, and $0$ for the baseline condition. We consider two complementary specifications. The \emph{difference model} pairs each perturbed observation with its baseline, so the baseline is absorbed into the difference and $d_i \in \{+1, -1\}$:
\begin{equation}
\Delta_i = \beta_{\text{pert}} \cdot d_i + \varepsilon_i, \quad \Delta_i = y_{\text{perturbed},i} - y_{\text{baseline},i}
\end{equation}
This yields a coefficient $\hat{\beta}_{\text{pert}}$ that estimates the directional shift associated with the perturbation.
We also consider a complementary \emph{level model} that includes the original-prompt output as a covariate (Appendix~\ref{app:level_model}); in our experiments, both specifications yield nearly identical estimates.
The effect magnitude is $|\hat{\beta}_{\text{pert}}|$; the direction is assessed through a one-sided hypothesis test on the sign of $\hat{\beta}_{\text{pert}}$ (Section~\ref{sec:testing}).

The limitation of MI, $\phi$, and flip rate is not that they aggregate (the mean of per-sample JSD values is an aggregate too) but that they discretize the output probabilities at the decision boundary, so they are lossy.
A shift from $P_i(\text{yes}) = 0.95$ to $Q_i(\text{yes}) = 0.85$, say, is a real change in the model's output distribution. 
But it does not cross the decision boundary, and so flip rate is blind to it.
Per-sample distributional metrics (JSD, KL) operate on the probabilities directly and so register every shift.

\vspace{-.5em}
\subsection{Statistical Testing}
\label{sec:testing}
\vspace{-.5em}
We want to establish if differences between changes in outputs resulting from targeted perturbations meaningfully exceed what we see under baseline perturbations. 

\paragraph{Paired $t$-test for per-sample metrics.} For JSD and KL, each prompt yields a pair comprising divergence values (under the targeted perturbation and the baseline, respectively). 
Our null hypothesis is that the targeted perturbation produces no larger effects than the baseline:
\begin{equation}
H_0: \mathbb{E}[\Delta_i] \leq 0, \quad H_1: \mathbb{E}[\Delta_i] > 0, \quad \Delta_i = d(p_i, p'_i) - d(p_i, p''_i)
\end{equation}
where $d$ is the per-sample JSD or KL, $p'_i$ is the targeted perturbation, and $p''_i$ is the baseline perturbation. We test this with a one-sided paired $t$-test on per-sample differences $\Delta_i$.

\paragraph{Bootstrap test for aggregate metrics.} For aggregate metrics (MI, $\phi$, flip rate), we obtain a single value per dataset rather than per-sample observations. We therefore adopt a nonparametric bootstrap: We resample the $N$ prompt pairs with replacement $B = 1{,}000$ times, compute the metric for both targeted and baseline conditions in each sample, and form the bootstrap distribution of the difference $\hat{\delta}^* = \hat{\Psi}^*_{\text{targeted}} - \hat{\Psi}^*_{\text{baseline}}$. The one-sided $p$-value is the proportion of bootstrap replicates inconsistent with the directional alternative: $\Pr(\hat{\delta}^* \le 0)$ for flip rate (where $H_1$: targeted~$>$~baseline), and $\Pr(\hat{\delta}^* \ge 0)$ for MI and $\phi$ (where $H_1$: targeted~$<$~baseline, since higher MI/$\phi$ indicates more agreement, i.e., less perturbation effect). Each bootstrap sample draws the prompt pairs once and evaluates both conditions on that same draw, preserving their pairing; Appendix~\ref{app:bootstrap} provides details.

\paragraph{Regression test.} When there is a prior directional hypothesis about how the perturbation should shift model outputs, we test the sign of the regression coefficient via a one-sided $t$-test: $H_0{:}\ \beta_{\text{pert}} \geq 0$ against $H_1{:}\ \beta_{\text{pert}} < 0$ (or the reverse). The divergence-based tests above ask whether the perturbation produces more \emph{change} (unsigned magnitude) than the baseline; the regression test additionally asks whether that change goes in the hypothesized \emph{direction}.


These tests quantify an average effect over particular textual versions realized by the targeted perturbation and the baseline, rather than the effect of the latent variable itself (Section~\ref{section:problem}).

\vspace{-.5em} 
\section{Experiments}
\vspace{-.75em} 

We assess whether the proposed controlled approach changes conclusions in practice. We revisit a medical triage study that claims significant effects of gender and language style perturbations (Section~\ref{section:revisit-medperturb}), validate the method through sanity checks and simulations (Section~\ref{section:sanity-checks}), apply it to occupation classification where bias is well-documented and expected (Section~\ref{section:bias}), and finally re-examine an audit of demographic bias in societal decisions (Appendix~\ref{app:discrimeval}).

\vspace{-.75em}
\subsection{Revisiting MedPerturb with a Baseline}
\label{section:revisit-medperturb}
\vspace{-.5em}

{\tt MedPerturb} \citep{gourabathina_medperturb_2025} is a benchmark recently introduced to assess the robustness of clinical LLMs to controlled perturbations made to factors of interest. 
We revisit their first case study, which focused on LLM sensitivity to gender and style perturbations. 
Using the MI metric defined above, the authors report that gender and style
perturbations influence LLMs more than human experts, 
suggesting LLMs are sensitive to these attributes when making treatment decisions. 
We revisit this analysis with perturbation baselines.  

\paragraph{Sensitivity, not appropriateness.}
It is sometimes appropriate for a model to (partially) rely on demographic factors.
Our framework does not offer judgment as to whether such reliance is appropriate; it assesses only if a model is \emph{especially} sensitive to an attribute, beyond baseline general sensitivity.
For the specific datasets we revisit, moreover, the manipulated attributes should not affect predictions: {\tt MedPerturb} \citep{gourabathina_medperturb_2025} performs its gender perturbations on gender attributes that should not affect clinical decision making, and prescription of pain medication should not depend on the race of a patient if all else is held equal, i.e., counterfactually (Q-Pain; \citealt{loge_q-pain_2021}, Appendices~\ref{app:qpain_replication} and~\ref{app:qpain_regression}).

\paragraph{Setup.}

We follow the {\tt MedPerturb} experimental setup as closely as possible. 
The dataset combines 100 clinical vignettes from two sources: OncQA (GPT-4-generated cancer patient summaries) and r/AskDocs (Reddit health questions answered by clinicians). Questions involving gender-specific conditions (e.g., ovarian or prostate cancer) are filtered out.
For each clinical vignette, the model answers three binary triage questions covering self-management at home (MANAGE), clinic or ED visit (VISIT), and resource allocation such as labs or specialist referrals (RESOURCE).
We use the same targeted perturbations from MedPerturb:

\vspace{-.25em}
\begin{tcolorbox}[colback=gray!10, colframe=black!50, boxrule=0.25pt,
  arc=2pt, top=2pt, bottom=2pt, left=4pt, right=4pt]
\footnotesize
\renewcommand{\arraystretch}{1.0}
\setlength{\tabcolsep}{4pt}
\begin{tabular}{@{}p{2.8cm}p{8.6cm}@{}}
  \textbf{Gender-swap}    & Swap all gender markers to the opposite gender \\[1pt]
  \textbf{Gender-remove}  & Remove all gender markers from the text \\[1pt]
  \textbf{Uncertain}      & Add uncertainty (e.g., ``maybe'', ``I think'', ``sort of'') \\[1pt]
  \textbf{Colorful}       & Add expressive language (e.g., exclamations, ``really'', ``very'')
\end{tabular}
\end{tcolorbox}
\vspace{0em}

For each perturbation type, we generate a paraphrase baseline (matched to the perturbation's token change percentage per sample) and a fixed sentence baseline (we prepend ``A family member came with me to the appointment.''). 
Each model generates three responses per prompt (seeds 0, 1, 42;  temperature 0.7) and the majority vote determines the binary answer, following the original {\tt MedPerturb} protocol; this affects only the aggregate metrics (MI, $\phi$, flip rate), not the per-sample metrics. 
We additionally extract next-token probabilities, from which the per-sample metrics (JSD, KL) are computed directly, without thresholding.

Again following the original {\tt MedPerturb} study, we evaluate {\tt Llama-3.1-8B-Instruct} and {\tt Llama-3.1-70B-Instruct}.\footnote{We use FP8 quantization to fit the 70B model on a single GPU.}
The main experiment tests 4 perturbation types $\times$ 3 tasks = 12 hypotheses per metric, baseline, and model. We apply Bonferroni correction within each group of 12 tests ($\alpha = 0.05/12 \approx 0.004$).

\paragraph{Do we need to account for the number of tokens changed?}
\label{sec:dose_response}
We have proposed to adjust matched paraphrases to account for the token change percentage of the targeted perturbation. To test if this control is necessary, we generate paraphrases at five token change levels (5\%, 10\%, 20\%, 40\%, 60\%) using GPT-5.2 and then evaluate {\tt Llama-3.1-8B-Instruct} on all triage questions, measuring flip rate and MI ($N = 100$).

\noindent\begin{minipage}[t]{0.52\columnwidth}
\vspace{0pt}
Figure~\ref{fig:dose_response} shows that flip rate increases with token change percentage for MANAGE (6\%~$\rightarrow$~14\%) and VISIT (5\%~$\rightarrow$~13\%); MI shows the corresponding decline (Appendix~\ref{app:dose_response_mi}).
(RESOURCE is flat due to class imbalance.)
This means a paraphrase that changes 5\% of tokens is a qualitatively different baseline than one changing 40\%. If a targeted perturbation changes 3\% of tokens but is compared against a baseline that edits 20\%, the targeted perturbation will look artificially clean by comparison (and the converse also holds). The adjusted paraphrase baseline avoids these pitfalls by matching the perturbation's token change magnitude.
\end{minipage}%
\hfill
\begin{minipage}[t]{0.45\columnwidth}
\vspace{0pt}
\centering
\includegraphics[width=\linewidth]{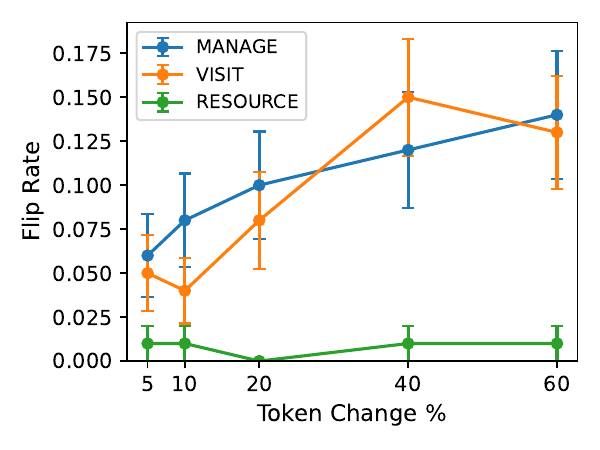}
\vspace{-1em}
\captionof{figure}{Flip rate increases with \% of tokens changed by benign paraphrasing.}
\label{fig:dose_response}
\end{minipage}

\begin{tcolorbox}[colback=blue!5, colframe=blue!40, boxrule=0.5pt, arc=2pt]
\textbf{Key Finding 1.} Model sensitivity to benign paraphrasing scales with the number of tokens changed, justifying magnitude-matched baselines for fair comparison.
\end{tcolorbox}

\paragraph{Results.}

Of 120 tests (4 perturbations $\times$ 3 tasks $\times$ 5 metrics $\times$ 2 models) against the adjusted paraphrase baseline (Table~\ref{tab:calibrated_results}), only 5 reach significance after Bonferroni correction---all for the colorful perturbation on the 8B model: flip rate on MANAGE, JSD on MANAGE and VISIT, and KL on MANAGE and VISIT. Four of these five are per-sample metrics; the only aggregate metric to reach significance anywhere in the table is the flip rate on MANAGE. 
Gender perturbations (swap and remove) show no significant effects on any task for either model: observed differences are small and inconsistent in sign, indicating that gender modifications do not produce effects distinguishable from paraphrasing. Full results with both baselines are reported in Appendix~\ref{app:full_medperturb}.

Appendix~\ref{app:self_consistency} additionally tests every perturbation against the model's own sampling variability, drawing 10 samples per prompt. There, 14 of the 24 (perturbation, task, model) cells exceed sampling noise, but 12 of those 14 do not survive the paraphrase baseline---controlling for output stochasticity alone is not sufficient.

\begin{tcolorbox}[colback=blue!5, colframe=blue!40, boxrule=0.5pt, arc=2pt]
\textbf{Key Finding 2.} Gender perturbation effects are statistically indistinguishable from paraphrasing (adjusting for number of tokens edited); only the ``colorful'' style perturbation on the smaller model shows significant effects.
\end{tcolorbox}

\vspace{-.75em}
\subsection{Sanity checks for the proposed testing framework}
\label{section:sanity-checks}
\vspace{-.75em}

While we believe that there is a first principles argument for accounting for sensitivity to benign counterfactual perturbations, the absence of ground truth (i.e., the true sensitivity to ``gender'' or similar targeted counterfactual perturbation) makes it difficult to establish the case for this empirically. 
Here we offer sanity checks and simulations as evidence to support the use of hypothesis testing against baseline perturbations in counterfactual analyses.


\paragraph{Recall check (positive control).} One risk is that the proposed framework is too strict, i.e., will not reveal meaningful sensitivities to targeted perturbations even where they exist.
Here we ascertain whether the procedure identifies shifts for what we would assume to be a very strong perturbation effect.
Specifically, we again gender-swap terms as our intervention, but then ask: ``Is this patient male?''.
We expect the resulting shift in prediction to be (much) larger than that observed under benign perturbations.

\begin{wraptable}{l}{0.5\columnwidth}
\vspace{0em}
\centering
\footnotesize
\begin{tabular}{l|ccccc}
\toprule
Check & MI & $\phi$ & FR & JSD & KL \\
\midrule
Recall & \textcolor{red}{\textbf{-.345}} & \textcolor{red}{\textbf{-1.67}} & \textcolor{red}{\textbf{.840}} & \textcolor{red}{\textbf{.701}} & \textcolor{red}{\textbf{7.67}} \\
Precision & .128 & .042 & -.020 & .004 & \textcolor{red}{\textbf{.016}} \\
\bottomrule
\end{tabular}
\caption{\raggedright Results for 8B against the adjusted paraphrase baseline ($N=100$). Values are differences. \textcolor{red}{\textbf{Red bold}}: $p<0.05$.}
\label{tab:sanity_adjusted}
\vspace{-1.2em}
\end{wraptable}

Table~\ref{tab:sanity_adjusted} shows that all metrics are significant ($p < 0.05$) for the 8B model.
Flip rates are 87\% (8B) and 96\% (70B), exceeding the adjusted paraphrase baseline by 84 and 96 percentage points respectively.
The proposed procedure is thus not so conservative that it fails to identify very strong effects.

\paragraph{Precision check (negative control).}
As a control, we use the same gender question but increment patient age by one year (e.g., 28$\rightarrow$29).
We assume that this should not affect inferences about patient gender.
Table~\ref{tab:sanity_adjusted} confirms this for the 8B model: KL is marginal at $p = 0.046$, just below the conventional 0.05 threshold; the other four metrics do not reach significance against the adjusted paraphrase baseline. Recall and precision check results for the 70B model and the fixed sentence baseline are in Appendix~\ref{app:sanity_fixed}. Results for the 70B model are consistent with results for the 8B model.

\begin{tcolorbox}[colback=blue!5, colframe=blue!40, boxrule=0.5pt, arc=2pt]
\textbf{Key Finding 3.} The testing framework detects strong effects and is well-controlled on a minimal perturbation that should, intuitively, have no effect.
\end{tcolorbox}

\paragraph{Metric Power Analysis via Simulation.}
\label{sec:power_simulation}

In the above experiments we do not know the true perturbation effect size ($\sigma_{\text{pert}}$) or baseline noise level  ($\sigma$)---we can only attempt to infer if a perturbation has an effect beyond the baseline. 
To assess if metrics have adequate power, we run Monte Carlo simulations, directly controlling $\sigma_{\text{pert}}$ and $\sigma$. 

Per-sample metrics (JSD, KL) are dramatically more powerful than population metrics (MI, $\phi$, flip rate): JSD offers strong detection at moderate effect sizes while MI, $\phi$, and flip rate remain near $\alpha = 0.05$. The gap is largest under class imbalance: in the RESOURCE condition on the 8B (99/100 positive), MI and $\phi$ produce zero detection at all effect sizes because the nearly constant answer distribution makes the contingency table degenerate; JSD and KL retain full sensitivity. For $\sigma > 0$, the null hypothesis is well-calibrated: the mean false positive rate at $\sigma_{\text{pert}} = 0$ is 0.043, with no single test exceeding 0.075. Full results in Appendix~\ref{app:power_curves}.










\vspace{-.8em}
\subsection{Directional Bias Detection on Bias-in-Bios}
\label{section:bias}
\vspace{-1em}
The preceding experiments test general sensitivity to targeted perturbations. 
But in some cases a directional hypothesis is warranted, e.g., we may \emph{a priori} suspect a model is biased against a particular group. 
We apply the regression framework to a (general domain) setting like this, and show that it can detect a real directional effect when it exists.  

\paragraph{Setup.}
Gender-occupation bias (the association of male with higher-status professions) is well-documented in society and in the text corpora on which language models are trained \citep{bolukbasi_man_2016, caliskan_semantics_2017}. It is therefore reasonable to expect that models encode this association. We test this directly using the bias-in-bios dataset \citep{de-arteaga_bias_2019}, which contains professional biographies labeled by occupation and gender.

Following \citet{marks_sparse_2025}, we construct a balanced subset of professor and nurse biographies featuring male and female genders, sampling $n = \min(\text{count of all 4 groups})$ per group. 
This yields 4{,}472 training and 1{,}724 test biographies. 
We present each biography to the model with a binary prompt (``What is this person's profession? A) Professor, B) Nurse'') and extract $\text{logit}_A - \text{logit}_B$ (positive values favor professor).
For each biography, we evaluate four text variants: (1)~the original biography; (2)~the gender-swapped biography (all gendered terms replaced with their opposites); (3)~an adjusted paraphrase; and (4)~the original biography prepended with a fixed sentence (``This profile has been viewed''). Gender swap is encoded as $\text{swap\_direction} = +1$ (swapped to female) or $-1$ (to male).

\begin{wraptable}{r}{0.32\columnwidth}
  \centering
  \vspace{-\intextsep}
  \footnotesize
  \setlength{\tabcolsep}{4pt}
  \renewcommand{\arraystretch}{1.1}
  \begin{tabular}{l S[table-format=+1.3] l}
    \toprule
    Measure & {Eff.} & {$p$} \\
    \midrule
    \multicolumn{3}{l}{\textit{Aggregate}} \\
    \quad FR            & +0.007 & $<.001$ \\
    \quad MI            & -0.053 & $<.001$ \\
    \quad $\phi$        & -0.015 & $<.001$ \\
    \multicolumn{3}{l}{\textit{Per-sample}} \\
    \quad JSD           & +0.005 & $10^{-15}$ \\
    \quad KL            & +0.022 & $10^{-7}$ \\
    \multicolumn{3}{l}{\textit{Regression}} \\
    \quad $\hat\beta_g$ & -0.485 & $\ll10^{-20}$ \\
    \bottomrule
  \end{tabular}
  \caption{\footnotesize Gender bias on \texttt{bias-in-bios} (Qwen3-32B)
    vs.\ baseline ($N{=}4{,}472$). Eff.: gender swap $-$ baseline;
    $\hat\beta_{\text{gender}}$ for regression. See Appendix~\ref{app:bib_test}.}
    \vspace{-2em}
  \label{table:bib_results}
\end{wraptable}

We evaluate {\tt Qwen3-8B} and {\tt Qwen3-32B} in non-thinking mode.
Our directional hypothesis is $H_1{:}\ \beta_{\text{gender}} < 0$ (i.e., swapping a biography to female will decrease the log-odds of professor). 
We apply both regression specifications from Section~\ref{sec:metrics} (run separately), with $d_i = \text{swap\_direction}_i$ and $\beta_{\text{pert}} = \beta_{\text{gender}}$. 
We train and test on disjoint splits.
For the level model, we use clustered SEs to account for within-biography correlation across the baseline and the perturbed rows. 

\paragraph{Results.}
Table~\ref{table:bib_results} reports results against the paraphrase baseline. 
All metrics detect a gender effect, but with varying magnitude and significance.
A $+0.7\%$ excess flip rate for {\tt Qwen3-32B} is significant but practically negligible; one might conclude the model is essentially unbiased. Per-sample distributional metrics confirm statistically measurable effects ($p \approx 10^{-7}$), but with small magnitudes (JSD $\approx 0.005$ on a $[0,1]$ scale), offering limited interpretive value. 
The regression coefficient reveals a real directional shift: $\hat{\beta}_{\text{gender}} \approx -0.5$ implies that swapping a biography to female decreases the log-odds of professor by about half a unit. And the sign and magnitude are directly interpretable.

The effect operates almost entirely below the decision boundary: logits have standard deviations of $\sim$15--20, so a shift of 0.5 rarely flips the binary prediction, explaining the near-zero flip rates. Per-sample metrics detect the distributional shift; regression characterizes it. Aggregate metrics, while significant here, lose significance entirely for the 8B model on the smaller test split (Appendix~\ref{app:bib_test}), consistent with the power gap identified in the simulation.

{\tt Qwen3-8B} shows slightly larger $|\hat{\beta}_{\text{gender}}|$ than {\tt Qwen3-32B} ($\approx 0.52$ vs.\ $\approx 0.49$), consistent with the above experiments in that larger models show lower sensitivity to perturbations.

We also apply the regression framework to race bias in pain management using the Q-Pain dataset \citep{loge_q-pain_2021}. Most regressions produce null results (Appendix~\ref{app:qpain_regression}). 
This again underscores the need to compare targeted counterfactuals to baselines.

\vspace{-.2em}
\begin{tcolorbox}[colback=blue!5, colframe=blue!40, boxrule=0.5pt, arc=2pt]
\textbf{Key Finding 4.} Per-sample distributional metrics detect effects; regression characterizes their direction and magnitude; aggregate metrics may miss them. 
\end{tcolorbox}

\paragraph{DiscrimEval.} Finally, we re-examine {\tt DiscrimEval} \citep{tamkin_evaluating_2023}, a benchmark designed to audit demographic bias in high-stakes societal decisions, for which a discrimination conclusion has already been published. Across its 70 ``Explicit'' scenarios $\times$ 14 demographic contrasts---each varying gender, race, or age against a white, 60-year-old, male reference profile---raw effects on both Llama models are substantial and qualitatively match those reported on Claude 2.0 (up to $+1.00$ log-odds for {\tt race\_Black} on the 70B), but no contrast reaches Bonferroni significance against the adjusted paraphrase baseline (0/14): the demographic signal is statistically indistinguishable from the model's response to a same-magnitude paraphrase. This is a re-analysis on a different model, not a contradiction of the original result; full setup and per-contrast results are in Appendix~\ref{app:discrimeval}.

\vspace{-.75em}
\section{Related Work}
\vspace{-.75em}

\paragraph{Counterfactual Bias Evaluation.}
Counterfactual pair benchmarks, such as {\tt CrowS-Pairs} \citep{nangia_crows-pairs_2020}, {\tt StereoSet} \citep{nadeem_stereoset_2021}, {\tt WinoBias} \citep{zhao_gender_2018}, and {\tt Winogender} \citep{rudinger_gender_2018}, measure bias by comparing outputs on prompts that differ only in demographic attributes, tacitly assuming that any differences can then be attributed to corresponding attributes. 
Prior critiques have challenged benchmark validity on various grounds  \citep{blodgett_stereotyping_2021,kohankhaki_template-based_2026,vamvas_limits_2021}. 

\paragraph{Chain-of-Thought (CoT) Faithfulness.}
Counterfactual pairs have been used to assess if CoT reasoning is \emph{faithful} \citep{jacovi_towards_2020}. \citet{lanham_measuring_2023} use truncation, mistake injection, and paraphrasing as faithfulness tests on CoT tokens. \citet{turpin_language_2023} and \citet{chen_reasoning_2025} inject hints into prompts to see if these are verbalized. 

\paragraph{Prompt Sensitivity.}
Substantial work documents model sensitivity to meaning-preserving changes \citep{webson-pavlick-2022-prompt, sun2023evaluating, sclar_quantifying_2024,cao_worst_2024}, motivating our use of benign perturbations (paraphrasing) as a baseline. 

\paragraph{Coupled Counterfactual Generation.}
Recent work couples sampling noise across generation runs, eliminating output stochasticity as a source of difference \citep{ravfogel_gumbel_2025, chatzi_counterfactual_2025, benz_evaluation_2026}.
The two approaches are complementary---coupled sampling controls output stochasticity, our baselines control input confounding (see Appendix~\ref{app:self_consistency}).
\paragraph{Statistical Rigor in Interpretability.}
A broader call for methodological rigor in LLM evaluation is emerging. \citet{meloux_dead_2025} show that interpretability methods—probing, SAEs, circuit discovery—can produce plausible explanations even for randomly initialized networks, advocating for hypothesis testing against null models in general. 

\vspace{-.6em}
\section{Conclusions, Practical Guidance, and Limitations}
\vspace{-1em}

Counterfactual prompting is widely used to evaluate LLM behavior with respect to factors of interest. 
Past work has tended to assume that observed output changes owe to the semantics of the perturbation rather than to general LLM sensitivity. We proposed testing targeted perturbation effects against a paraphrase baseline (adjusting for edit lengths). 

Applied to {\tt MedPerturb} \citep{gourabathina_medperturb_2025}, e.g., only 5/120  tests reached significance against the adjusted baseline. We validate our framework through sanity checks that demonstrate its recall and precision. On {\tt Bias-in-Bios} \citep{de-arteaga_bias_2019}, we detected significant directional gender bias ($\hat{\beta}_{\text{gender}} \approx -0.5$). 
On {\tt DiscrimEval} \citep{tamkin_evaluating_2023}, no demographic contrast survived comparison to the same baseline, despite raw effects reaching one log-odds unit. 
Without quantifying with respect to a baseline, counterfactual evaluations may conflate targeted effects with general prompt sensitivity. 

\paragraph{Practical guidance.} \textbf{(1) Include a magnitude-matched baseline.} Paraphrase baselines must match the perturbation's token change percentage, since model sensitivity scales with this percentage. \textbf{(2) Use per-sample distributional metrics as primary evidence.} Per-sample distributional metrics (JSD, KL) are far more powerful than aggregate metrics (MI, $\phi$, flip rate). \textbf{(3) Use regression for directional hypotheses.} When the perturbation admits a directional hypothesis, regression estimates direction and magnitude; this is powerful when characterization of the effect matters. \textbf{(4) Test for significance.} Differences between perturbation and baseline should be statistically tested, not interpreted at face value.

\paragraph{Limitations.}
LLM-generated paraphrases may not perfectly preserve semantics which could inflate baseline effects.
In clinical texts, e.g., even small changes in wording can shift emphasis, perceived reliability of the patient, or plausible differential diagnoses; no rewording of a vignette is guaranteed to preserve meaning. 
When paraphrases do alter meaning, this induces real semantic effects, and so the reference distribution is larger than the surface-form sensitivity it is meant to capture.
Our tests then become conservative: A targeted effect must clear a higher bar to reach significance.
This is a further reason to read null results as indistinguishability from the reference and not (necessarily) the absence of an effect.
Our medical experiments also use small datasets, limiting statistical power.
Finally, the JSD and KL metrics require next-token logits, which are not meaningful for open-ended generation and not available for black-box LLMs (e.g., Claude).

\vspace{-.5em}
\section*{Acknowledgments}
\vspace{-.5em}
This work was supported by Coefficient Giving. 

\bibliography{references_zotero,references}
\bibliographystyle{colm2026_conference}

\appendix

\section{Level Model Regression Specification}
\label{app:level_model}
The \emph{level model} stacks baseline ($d_i = 0$) and perturbed ($d_i = \pm 1$) observations together and includes model outputs (logits) on the original prompt as a covariate $z_i$:
\begin{equation}
y_i = \beta_0 + \beta_1 z_i + \beta_{\text{pert}} \cdot d_i + \varepsilon_i
\end{equation}
By including $z_i$, the level model controls for prompt-level variation in model output, which may improve precision. In our experiments both specifications yield nearly identical $\hat{\beta}_{\text{pert}}$ estimates.


\section{Gender Swapping on MedQA}
\label{app:medqa}

To illustrate the problem in a minimal setup, we conduct preliminary experiments with two DeepSeek-R1 distilled models, one Qwen3-based (8B) and one Llama-based (70B).

We use MedQA \citep{jin_what_2020}, a multiple-choice medical question answering benchmark. 
We select MedQA because current models achieve high accuracy on this benchmark, reducing noise from random guessing.
Each question presents a clinical vignette and four answer options. We filter out questions where gender is clinically relevant---so any sensitivity to gender swaps is inappropriate---and questions unrelated to diagnosis, resulting in 919 samples.

As a targeted perturbation, we swap all gender-related terms  (e.g., ``he'' $\rightarrow$ ``she'', ``male'' $\rightarrow$ ``female''). For the adjusted paraphrase baseline, we paraphrase a randomly selected sentence in each query using GPT-5.2, matching the token change percentage of the corresponding gender perturbation.

We evaluate using both flip rate and JSD metrics with statistical hypothesis testing.

\paragraph{Results.}

\begin{table*}[h]
\centering
\begin{tabular}{l|ccc|ccc}
\toprule
Model & JSD (G) & JSD (B) & $p$ & FR (G) & FR (B) & $p$ \\
\midrule
\shortstack[l]{DeepSeek-R1-0528\\Qwen3-8B} & 0.101 & 0.096 & 0.285 & 14.9\% & 14.1\% & 0.272 \\
\shortstack[l]{DeepSeek-R1-Distill\\Llama-70B} & 0.041 & 0.032 & 0.016 & 7.4\% & 6.0\% & 0.038 \\
\bottomrule
\end{tabular}
\caption{Preliminary results on MedQA comparing gender perturbation (G) effects against the adjusted paraphrase baseline (B). JSD $p$-values are from one-sided paired $t$-tests ($H_1$: gender~$>$~baseline); flip rate (FR) $p$-values are from one-sided bootstrap resampling ($B = 10{,}000$). The names of both DeepSeek-R1 models are their HuggingFace names.}
\label{tab:preliminary_results}
\end{table*}

Table~\ref{tab:preliminary_results} presents our results comparing gender perturbation effects against the adjusted paraphrase baseline on MedQA.

For the 8B model, \emph{we observe no significant difference between gender perturbation and the adjusted baseline}. The mean JSD values are nearly identical (0.101 vs.\ 0.096, paired $t$-test $p = 0.285$), and flip rates show the same pattern (14.9\% vs.\ 14.1\%, bootstrap $p = 0.272$). This suggests that observed output changes from gender swapping do not exceed what would be expected from magnitude-matched paraphrasing.

For the 70B model, both metrics detect that gender perturbation produces significantly greater divergence than the baseline: paired $t$-test on JSD ($p = 0.016$, mean 0.041 vs.\ 0.032) and bootstrap on flip rate (7.4\% vs.\ 6.0\%, $p = 0.038$). The absolute differences remain small. This suggests that while gender perturbation does produce effects beyond paraphrasing for the larger model, the magnitude is modest---substantially smaller than what one might conclude from the raw perturbation results alone, without consideration of a baseline.

Notably, the larger 70B model shows lower overall sensitivity to both perturbation types, with JSD and flip rates roughly half those of the 8B model. This aligns with prior findings that larger models tend to be comparatively robust to prompt variations.

These preliminary results support our hypothesis that surgical perturbations like gender swapping may not produce effects reliably distinguishable from general prompt sensitivity, at least for smaller-scale interventions. 
Even when they do, as for the bigger model here, taking into account the baseline sensitivity seems important to appropriately interpret perturbation results. 

\section{Sensitivity to token change magnitude: MI Plot}
\label{app:dose_response_mi}

\begin{figure}[h]
    \centering
    \includegraphics[width=0.6\columnwidth]{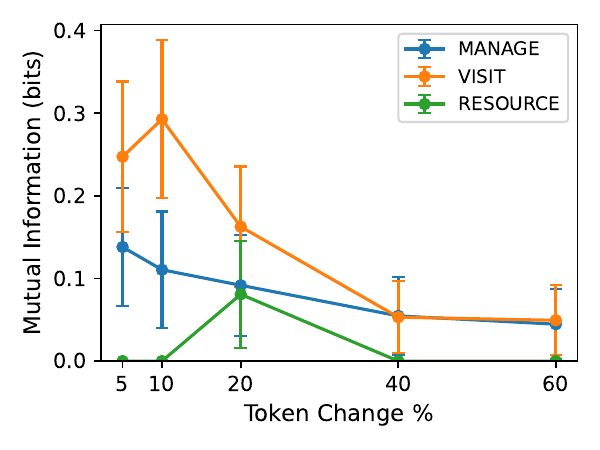}
    \caption{Mutual information between original and paraphrased responses decreases with token change percentage.}
    \label{fig:dose_response_mi}
\end{figure}

Figure~\ref{fig:dose_response_mi} shows the MI counterpart to the flip rate analysis in Figure~\ref{fig:dose_response}: MI between original and paraphrased responses decreases with token change percentage, indicating less consistency under larger benign perturbations.

\section{Full Results of MedPerturb Experiments}
\label{app:full_medperturb}

\begin{table*}[t]
\centering
\resizebox{\textwidth}{!}{%
\begin{tabular}{ll|ccccc|ccccc}
\toprule
& & \multicolumn{5}{c|}{Llama 3.1 8B} & \multicolumn{5}{c}{Llama 3.1 70B} \\
Perturbation & Task & MI & $\phi$ & FR & JSD & KL & MI & $\phi$ & FR & JSD & KL \\
\midrule
\multirow{3}{*}{Gender-swap}
    & MAN & .000 & .000 & .000 & -.003 & -.017 & .022 & .005 & .000 & .003 & .038 \\
    & VIS & -.028 & -.046 & .010 & -.001 & -.003 & .018 & .003 & .000 & -.001 & -.001 \\
    & RES & .000 & .000 & .000 & -.000 & -.001 & -.012 & -.094 & .010 & .006 & .030 \\
\midrule
\multirow{3}{*}{Gender-remove}
    & MAN & .050 & .089 & -.020 & -.001 & -.004 & .000 & .000 & .000 & .001 & .003 \\
    & VIS & .105 & .119 & -.020 & .002 & .010 & .068 & .056 & -.020 & -.001 & -.006 \\
    & RES & .081 & 1.00 & -.010 & .000 & .001 & .141 & .429 & -.020 & -.000 & -.003 \\
\midrule
\multirow{3}{*}{Uncertain}
    & MAN & -.022 & -.055 & .000 & -.003 & -.013 & -.020 & -.024 & .000 & .015 & .260 \\
    & VIS & -.091 & -.277 & .020 & .014 & .090 & .246 & .272 & -.100 & -.034 & -.235 \\
    & RES & -.000 & .010 & -.010 & -.003 & .009 & -.032 & .060 & -.030 & -.027 & -.131 \\
\midrule
\multirow{3}{*}{Colorful}
    & MAN & -.046 & -.181 & \textcolor{red}{\textbf{.170}} & \textcolor{red}{\textbf{.027}} & \textcolor{red}{\textbf{.109}} & -.113 & -.141 & .020 & .005 & .160 \\
    & VIS & -.200 & -.437 & .060 & \textcolor{red}{\textbf{.031}} & \textcolor{red}{\textbf{.165}} & -.045 & -.022 & .000 & -.008 & -.020 \\
    & RES & .000 & .000 & .000 & .004 & .042 & -.044 & -.234 & .010 & -.007 & -.025 \\
\bottomrule
\end{tabular}}
\caption{Main experiment results against the adjusted paraphrase baseline ($N = 100$). Values are observed differences (perturbation $-$ baseline); \textcolor{red}{\textbf{Red bold}}: perturbation produces significantly greater divergence than the baseline ($p < 0.05/12$, Bonferroni-corrected). Of 120 tests (4 perturbations $\times$ 3 tasks $\times$ 5 metrics $\times$ 2 models), only 5 reach significance, all for the colorful perturbation on the 8B model.}
\label{tab:calibrated_results}
\end{table*}

\paragraph{Adjusted paraphrase baseline.} Table~\ref{tab:calibrated_results} reports the main results against the adjusted paraphrase baseline. Of 120 tests across both models, only 5 reach significance after Bonferroni correction---all for the colorful perturbation on the 8B model. 
This perturbation increases the flip rate on MANAGE by 17 percentage points beyond the baseline ($p = 0.001$), and produces significant JSD and KL divergence on both MANAGE and VISIT.

Gender perturbations (swap and remove) show no significant effects on any task for either model. The observed differences are small and inconsistent in sign, indicating that gender modifications do not produce effects distinguishable from token-adjusted paraphrasing. This holds for all five metrics, including the per-sample logit-based measures (JSD, KL) that are sensitive to distributional shifts below the decision boundary.

The uncertain perturbation also fails to reach significance against the adjusted baseline for either model, despite producing notable raw effects on some tasks (e.g., uncertain MANAGE for 70B raises KL by $0.26$ over the adjusted baseline, $p = .016$, which does not survive Bonferroni correction).

Several cells in Table~\ref{tab:calibrated_results} show exact zeros for MI, $\phi$, and flip rate---particularly for RESOURCE and MANAGE. 
All three triage tasks exhibit class imbalance: on the original (unperturbed) prompts, the 8B model answers ``yes'' on 99\% of RESOURCE cases, 89\% of VISIT cases, and only 10\% of MANAGE cases. When the class imbalance is extreme and the perturbation is small, neither the perturbation nor the baseline flips any answers, making per-population metrics identically zero. The per-sample metrics (JSD, KL) are not affected by this imbalance because they operate on probability distributions rather than discretized answers. In Section~\ref{sec:power_simulation} we use simulations to confirm that per-population metrics are fundamentally limited under class imbalance.

The 70B model shows no significant effects for any perturbation type against the adjusted baseline (0/60), consistent with larger models being more robust to prompt perturbations.

\paragraph{Fixed sentence baseline.}

\begin{table*}[t]
\centering
\resizebox{\textwidth}{!}{%
\begin{tabular}{ll|ccccc|ccccc}
\toprule
& & \multicolumn{5}{c|}{Llama 3.1 8B} & \multicolumn{5}{c}{Llama 3.1 70B} \\
Perturbation & Task & MI & $\phi$ & FR & JSD & KL & MI & $\phi$ & FR & JSD & KL \\
\midrule
\multirow{3}{*}{Gender-swap}
    & MAN & -.130 & -.127 & .020 & -.001 & -.005 & -.032 & -.032 & .010 & .005 & .042 \\
    & VIS & .000 & .000 & .000 & -.002 & -.009 & .018 & .003 & .000 & -.006 & -.027 \\
    & RES & .000 & .000 & .000 & -.000 & -.001 & -.120 & -.438 & .030 & .003 & .019 \\
\midrule
\multirow{3}{*}{Gender-remove}
    & MAN & -.025 & .027 & -.010 & -.001 & -.004 & .044 & .006 & .000 & -.000 & -.001 \\
    & VIS & .056 & .058 & -.010 & -.001 & -.001 & .000 & .000 & .000 & -.006 & -.027 \\
    & RES & .081 & 1.00 & -.010 & -.001 & -.002 & .000 & .000 & .000 & -.003 & -.014 \\
\midrule
\multirow{3}{*}{Uncertain}
    & MAN & \textcolor{red}{\textbf{-.311}} & \textcolor{red}{\textbf{-.536}} & \textcolor{red}{\textbf{.120}} & \textcolor{red}{\textbf{.021}} & \textcolor{red}{\textbf{.089}} & \textcolor{red}{\textbf{-.321}} & \textcolor{red}{\textbf{-.322}} & \textcolor{red}{\textbf{.080}} & \textcolor{red}{\textbf{.051}} & \textcolor{red}{\textbf{.505}} \\
    & VIS & \textcolor{red}{\textbf{-.265}} & \textcolor{red}{\textbf{-.583}} & .080 & \textcolor{red}{\textbf{.045}} & \textcolor{red}{\textbf{.216}} & -.094 & -.064 & .020 & \textcolor{red}{\textbf{.025}} & .127 \\
    & RES & .000 & .000 & .000 & \textcolor{red}{\textbf{.012}} & \textcolor{red}{\textbf{.069}} & -.108 & -.344 & .020 & .014 & \textcolor{red}{\textbf{.155}} \\
\midrule
\multirow{3}{*}{Colorful}
    & MAN & \textcolor{red}{\textbf{-.339}} & \textcolor{red}{\textbf{-.653}} & \textcolor{red}{\textbf{.260}} & \textcolor{red}{\textbf{.039}} & \textcolor{red}{\textbf{.158}} & \textcolor{red}{\textbf{-.377}} & \textcolor{red}{\textbf{-.407}} & \textcolor{red}{\textbf{.100}} & \textcolor{red}{\textbf{.038}} & \textcolor{red}{\textbf{.326}} \\
    & VIS & \textcolor{red}{\textbf{-.255}} & \textcolor{red}{\textbf{-.546}} & .090 & \textcolor{red}{\textbf{.054}} & \textcolor{red}{\textbf{.264}} & -.210 & -.162 & .050 & \textcolor{red}{\textbf{.037}} & \textcolor{red}{\textbf{.212}} \\
    & RES & .000 & .000 & .000 & \textcolor{red}{\textbf{.014}} & .083 & -.173 & -.737 & .050 & .019 & .176 \\
\bottomrule
\end{tabular}}
\caption{Main experiment results against the fixed sentence baseline ($N = 100$). Values are observed differences (perturbation $-$ baseline); \textcolor{red}{\textbf{Red bold}}: perturbation produces significantly greater divergence than the baseline ($p < 0.05/12$, Bonferroni-corrected, one-sided). Style perturbations (uncertain, colorful) substantially exceed this minimal baseline, while gender perturbations largely do not.}
\label{tab:neutral_results}
\end{table*}

Table~\ref{tab:neutral_results} shows results against the fixed sentence baseline. Against this simpler baseline, a strikingly different pattern emerges. Style perturbations (uncertain and colorful) now show significance: colorful MANAGE produces 26 percentage points of excess flips for 8B and 10 for 70B ($p < 0.001$), with large JSD and KL effects across multiple tasks. Uncertain language shows a similar pattern, particularly on MANAGE. However, this comparison is confounded by a mismatch in the number of tokens changed: The uncertain and colorful perturbations change substantially more tokens than the fixed sentence. Our analysis (Section~\ref{sec:dose_response}) shows that model sensitivity scales with the fraction of tokens changed, so much of the observed significance likely reflects this size difference rather than the semantic content of the perturbation.

Gender perturbations, by contrast, remain non-significant even against this simpler baseline for both models on all three tasks across all five metrics.

The contrast between the two baselines is informative. Style perturbations produce effects that exceed the fixed sentence baseline but not the adjusted paraphrase baseline (for the 70B model) or only partially (for the 8B model). That is, these perturbations change model outputs more than a minimal prompt modification does, but not more than a semantically neutral paraphrase of comparable magnitude. Gender perturbations fail to exceed either baseline, providing strong evidence that their observed effects in prior work reflect general prompt sensitivity rather than gender-specific model behavior.

\section{Recall and Precision Checks: Full Results}
\label{app:sanity_fixed}

Table~\ref{tab:sanity_adjusted_full} reports the recall and precision checks against the adjusted paraphrase baseline for both models. The 70B model shows all metrics significant for the recall check and little effect for the precision check.

\begin{table*}[h]
\centering
\small
\begin{tabular}{l|ccccc|ccccc}
\toprule
& \multicolumn{5}{c|}{Llama 3.1 8B} & \multicolumn{5}{c}{Llama 3.1 70B} \\
Check & MI & $\phi$ & FR & JSD & KL & MI & $\phi$ & FR & JSD & KL \\
\midrule
Recall
    & \textcolor{red}{\textbf{-.345}} & \textcolor{red}{\textbf{-1.67}} & \textcolor{red}{\textbf{.840}} & \textcolor{red}{\textbf{.701}} & \textcolor{red}{\textbf{7.67}} & \textcolor{red}{\textbf{-.217}} & \textcolor{red}{\textbf{-1.92}} & \textcolor{red}{\textbf{.960}} & \textcolor{red}{\textbf{.948}} & \textcolor{red}{\textbf{19.55}} \\
Precision
    & .128 & .042 & -.020 & .004 & \textcolor{red}{\textbf{.016}} & .000 & .000 & .000 & .000 & -.001 \\
\bottomrule
\end{tabular}
\caption{Recall and precision checks against the adjusted paraphrase baseline ($N = 100$), both models. Values are differences (perturbation $-$ baseline). \textcolor{red}{\textbf{Red bold}}: perturbation produces significantly greater divergence than the baseline (one-sided $p < 0.05$).}
\label{tab:sanity_adjusted_full}
\end{table*}

Table~\ref{tab:sanity_fixed} reports the recall and precision checks against the fixed sentence baseline. The recall check yields qualitatively identical results to the adjusted baseline (Table~\ref{tab:sanity_adjusted_full}): all five metrics significant for both models. For the precision check, age~$+1$ does not produce significantly greater divergence than the fixed-sentence baseline on any of the five metrics for the 8B model; the observed differences are in fact in the opposite direction (the fixed-sentence baseline shifts more than age~$+1$ on JSD, KL, and FR), confirming that incrementing age by one year does not have more effects than prepending an irrelevant sentence. The 70B model shows no significant effects.

\begin{table*}[h]
\centering
\small
\begin{tabular}{l|ccccc|ccccc}
\toprule
& \multicolumn{5}{c|}{Llama 3.1 8B} & \multicolumn{5}{c}{Llama 3.1 70B} \\
Check & MI & $\phi$ & FR & JSD & KL & MI & $\phi$ & FR & JSD & KL \\
\midrule
Recall
    & \textcolor{red}{\textbf{-.214}} & \textcolor{red}{\textbf{-1.60}} & \textcolor{red}{\textbf{.810}} & \textcolor{red}{\textbf{.689}} & \textcolor{red}{\textbf{7.62}} & \textcolor{red}{\textbf{-.217}} & \textcolor{red}{\textbf{-1.92}} & \textcolor{red}{\textbf{.960}} & \textcolor{red}{\textbf{.943}} & \textcolor{red}{\textbf{19.54}} \\
Precision
    & .325 & .127 & -.060 & -.016 & -.066 & .000 & .000 & .000 & -.005 & -.018 \\
\bottomrule
\end{tabular}
\caption{Recall and precision checks against the fixed sentence baseline ($N = 100$). Same format as Table~\ref{tab:sanity_adjusted_full}. \textcolor{red}{\textbf{Red bold}}: perturbation produces significantly greater divergence than the baseline (one-sided $p < 0.05$).}
\label{tab:sanity_fixed}
\end{table*}

\section{Power Analysis via Simulation}
\label{app:power_curves}

\paragraph{Model specification.} For each (question, model) condition, we take the model's real output logits $z_i$ from the original prompts as the starting point, where $z_i = \log(p_i^{\text{Yes}} / p_i^{\text{No}})$. We simulate perturbation and baseline arms by adding Gaussian noise in logit space:
\begin{align}
\text{logit}_{\text{pert},i} &= z_i + \epsilon_{\text{pert},i} + \epsilon_i, \quad
\text{logit}_{\text{base},i} = z_i + \epsilon_i'
\end{align}
where $\epsilon_{\text{pert},i} \sim \mathcal{N}(0, \sigma_{\text{pert}}^2)$ is the perturbation-specific effect and $\epsilon_i, \epsilon_i' \sim \mathcal{N}(0, \sigma^2)$ are independent general baseline noise draws. Logits are converted to probabilities via the sigmoid function. For per-sample metrics (JSD, KL), we compare probability distributions directly. For per-population metrics (MI, $\phi$, flip rate), we draw discrete binary answers $y \sim \text{Bernoulli}(p)$. Under $H_0$ ($\sigma_{\text{pert}} = 0$), both arms are symmetric. We sweep $\sigma_{\text{pert}}$ from 0 to 3.0 and $\sigma \in \{0, 0.25, 0.5, 1.0\}$, running 1{,}000 simulations per combination across all six conditions.

\begin{figure}[h]
    \centering
    \includegraphics[width=\columnwidth]{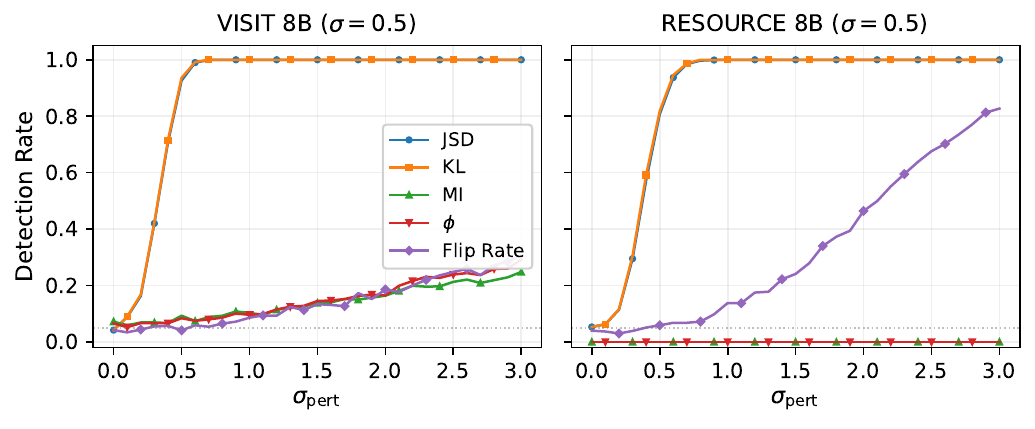}
    \caption{Power curves at $\sigma = 0.5$ for two 8B conditions. \textbf{Left (VISIT):} Per-sample metrics (JSD, KL) reach near-perfect detection while per-population metrics (MI, $\phi$, flip rate) remain near $\alpha = 0.05$. \textbf{Right (RESOURCE):} Under extreme class imbalance (99/100 positive), MI and $\phi$ are completely degenerate, while JSD and KL retain full sensitivity.}
    \label{fig:power_combined}
\end{figure}

\paragraph{Summary.} Figure~\ref{fig:power_combined} shows power curves at $\sigma = 0.5$ for two representative conditions. Per-sample metrics (JSD, KL) are dramatically more powerful than per-population metrics (MI, $\phi$, flip rate), which become degenerate under class imbalance. Studies relying solely on aggregate metrics may systematically fail to detect real effects. The null hypothesis is well-calibrated: across all six conditions, metrics, and $\sigma > 0$ values, the mean false positive rate at $\sigma_{\text{pert}} = 0$ is 0.043 (nominal: 0.05), with no single test exceeding 0.075.

\begin{figure}[h]
    \centering
    \includegraphics[width=\textwidth]{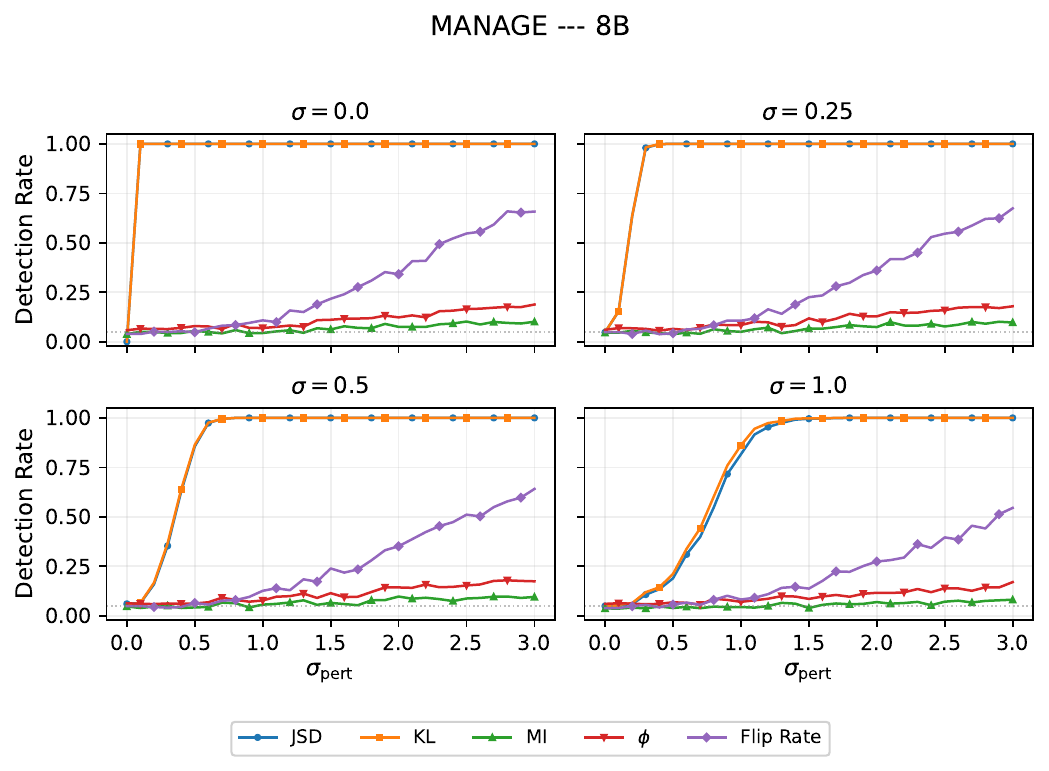}
    \caption{Power curves for MANAGE --- 8B.}
    \label{fig:power_manage_8b}
\end{figure}

\begin{figure}[h]
    \centering
    \includegraphics[width=\textwidth]{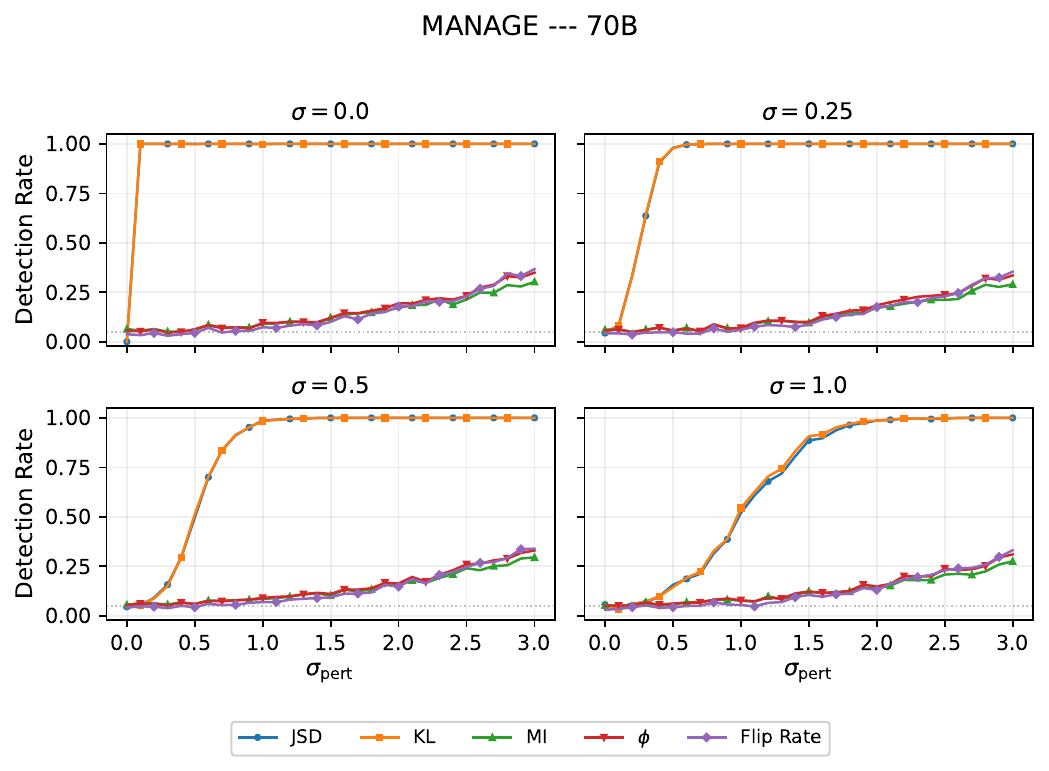}
    \caption{Power curves for MANAGE --- 70B.}
    \label{fig:power_manage_70b}
\end{figure}

\begin{figure}[h]
    \centering
    \includegraphics[width=\textwidth]{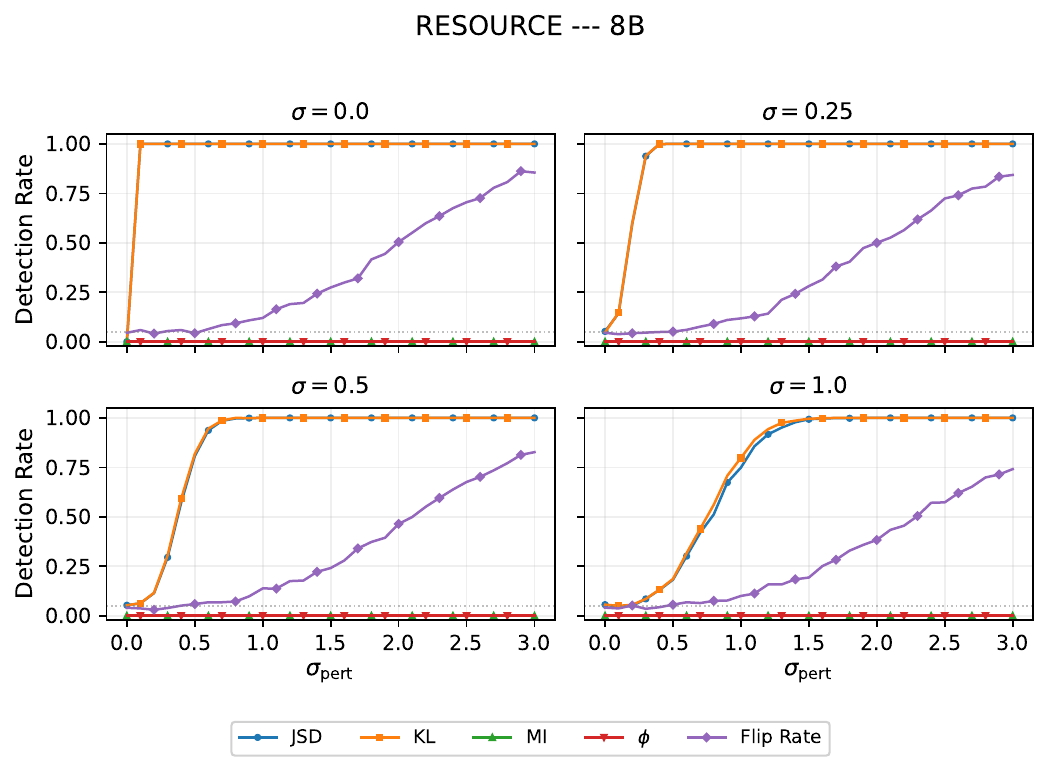}
    \caption{Power curves for RESOURCE --- 8B. MI and $\phi$ show zero detection across all $\sigma_{\text{pert}}$ values due to extreme class imbalance (99/100 positive), which makes contingency-table-based metrics degenerate.}
    \label{fig:power_resource_8b}
\end{figure}

\begin{figure}[h]
    \centering
    \includegraphics[width=\textwidth]{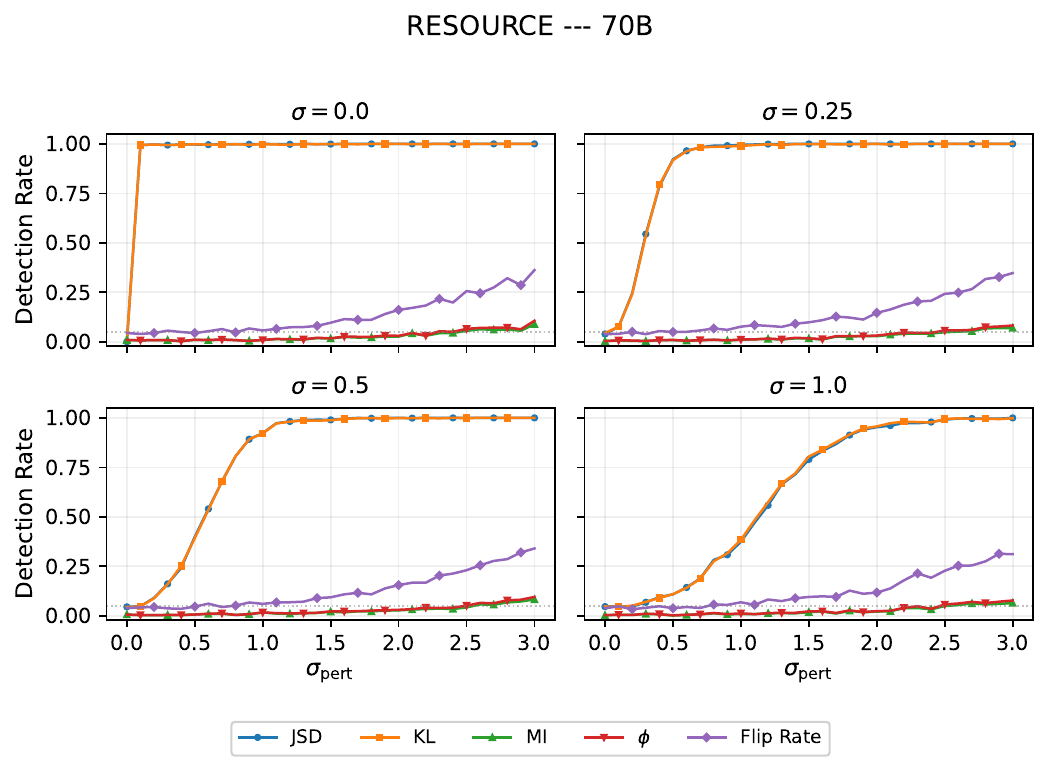}
    \caption{Power curves for RESOURCE --- 70B. MI and $\phi$ are near-powerless (detection $\leq 0.11$) rather than identically zero as on the 8B; JSD/KL power is reduced by highly confident predictions.}
    \label{fig:power_resource_70b}
\end{figure}

\begin{figure}[h]
    \centering
    \includegraphics[width=\textwidth]{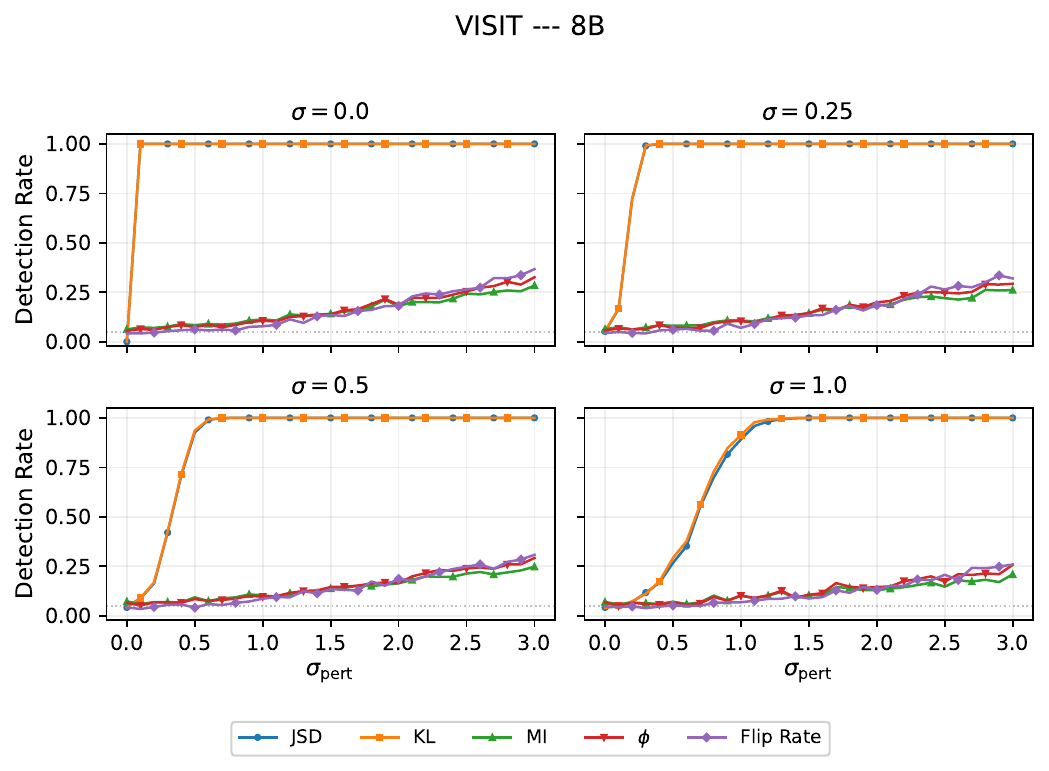}
    \caption{Power curves for VISIT --- 8B.}
    \label{fig:power_visit_8b}
\end{figure}

\begin{figure}[h]
    \centering
    \includegraphics[width=\textwidth]{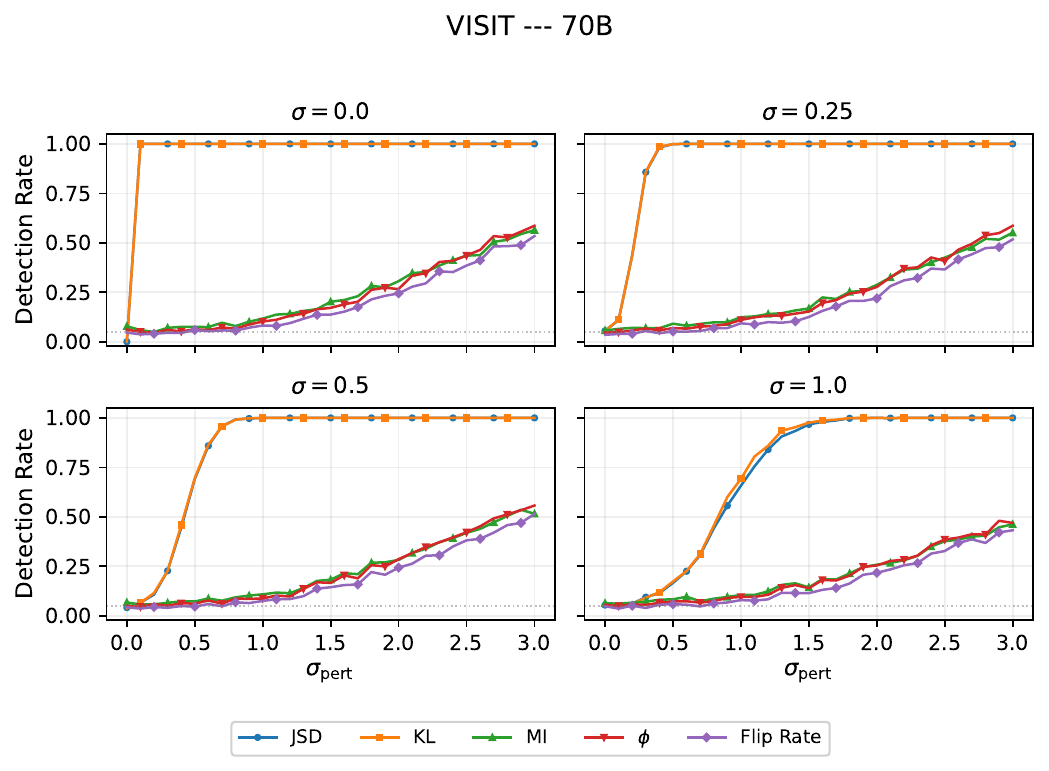}
    \caption{Power curves for VISIT --- 70B.}
    \label{fig:power_visit_70b}
\end{figure}

\paragraph{Per-condition results.} Figures~\ref{fig:power_manage_8b}--\ref{fig:power_visit_70b} show the full power curves for all six (question, model) conditions across four baseline noise levels. The pattern is consistent: per-sample metrics achieve high power at moderate $\sigma_{\text{pert}}$, while per-population metrics remain near $\alpha = 0.05$. At $\sigma = 0$, per-sample metrics show a step function (degenerate case where the baseline arm has zero noise). The 70B conditions require larger $\sigma_{\text{pert}}$ to achieve equivalent power due to more confident predictions.

\section{Bias-in-Bios: Full Results}
\label{app:bib_test}

Table~\ref{tab:bib_results_train} reports the full train-split results for both models against the adjusted paraphrase baseline. The main text (Table~\ref{table:bib_results}) reports only Qwen3-32B; here we include Qwen3-8B as well. Both models show consistent patterns: all metrics are significant, with per-sample and regression measures providing the clearest signal.

\begin{table}[h]
\centering
\small
\begin{tabular}{l|cc|cc}
\toprule
& \multicolumn{2}{c|}{Qwen3-8B} & \multicolumn{2}{c}{Qwen3-32B} \\
Measure & Effect & $p$ & Effect & $p$ \\
\midrule
\multicolumn{5}{l}{\textit{Aggregate}} \\
\midrule
Flip rate & $+0.006$ & ${<}\,0.001$ & $+0.007$ & ${<}\,0.001$ \\
MI & $-0.045$ & ${<}\,0.001$ & $-0.053$ & ${<}\,0.001$ \\
$\phi$ & $-0.012$ & ${<}\,0.001$ & $-0.015$ & ${<}\,0.001$ \\
\midrule
\multicolumn{5}{l}{\textit{Per-sample}} \\
\midrule
JSD & $+0.004$ & $4.1 \times 10^{-10}$ & $+0.005$ & $10^{-15}$ \\
KL & $+0.022$ & $3.9 \times 10^{-9}$ & $+0.022$ & $2.4 \times 10^{-7}$ \\
\midrule
\multicolumn{5}{l}{\textit{Regression}} \\
\midrule
$\hat{\beta}_{\text{gender}}$ & $-0.524$ & $\ll 10^{-20}$ & $-0.485$ & $\ll 10^{-20}$ \\
\bottomrule
\end{tabular}
\caption{Gender bias on bias-in-bios against the adjusted paraphrase baseline (train split, $N{=}4{,}472$). ``Effect'' is the difference (gender swap $-$ baseline) for aggregate/per-sample measures, or $\hat{\beta}_{\text{gender}}$ for regression. All $p$-values are one-sided. Bootstrap $p$-values reported as ${<}\,0.001$ when no replicate falls on the wrong side of zero ($B{=}1{,}000$).}
\label{tab:bib_results_train}
\end{table}

Table~\ref{tab:bib_results_test} reports the same metrics on the test split ($N{=}1{,}724$). Per-sample metrics (JSD, KL) remain significant across both models. Aggregate metrics, however, lose significance for Qwen3-8B (flip rate $p = 0.093$, MI $p = 0.078$), illustrating the power gap identified in the simulation study (Section~\ref{sec:power_simulation}): the same real effect that is detectable with per-sample metrics becomes invisible to aggregate metrics at smaller sample sizes.

\begin{table}[h]
\centering
\small
\begin{tabular}{l|cc|cc}
\toprule
& \multicolumn{2}{c|}{Qwen3-8B} & \multicolumn{2}{c}{Qwen3-32B} \\
Measure & Effect & $p$ & Effect & $p$ \\
\midrule
\multicolumn{5}{l}{\textit{Aggregate}} \\
\midrule
Flip rate & $+0.003$ & $0.093$ & $+0.008$ & $0.001$ \\
MI & $-0.022$ & $0.078$ & $-0.051$ & ${<}\,0.001$ \\
$\phi$ & $-0.006$ & $0.078$ & $-0.015$ & ${<}\,0.001$ \\
\midrule
\multicolumn{5}{l}{\textit{Per-sample}} \\
\midrule
JSD & $+0.004$ & $2.2 \times 10^{-5}$ & $+0.006$ & $6.6 \times 10^{-8}$ \\
KL & $+0.022$ & $9.8 \times 10^{-5}$ & $+0.033$ & $1.4 \times 10^{-6}$ \\
\midrule
\multicolumn{5}{l}{\textit{Regression}} \\
\midrule
$\hat{\beta}_{\text{gender}}$ & $-0.551$ & $\ll 10^{-20}$ & $-0.468$ & $\ll 10^{-20}$ \\
\bottomrule
\end{tabular}
\caption{Gender bias on bias-in-bios against the adjusted paraphrase baseline (test split, $N{=}1{,}724$). Same format as Table~\ref{tab:bib_results_train}. Aggregate metrics lose significance for Qwen3-8B while per-sample and regression measures remain highly significant.}
\label{tab:bib_results_test}
\end{table}

Tables~\ref{tab:bib_results_fixed_train} and~\ref{tab:bib_results_fixed_test} repeat the analysis against the fixed sentence baseline. The per-sample metrics and the regression remain highly significant for both models on both splits ($\hat{\beta}_{\text{gender}}$ between $-0.49$ and $-0.57$, $p \ll 10^{-20}$), matching the adjusted-paraphrase results. The aggregate metrics are again the fragile ones: non-significant for Qwen3-8B on the test split ($p \geq 0.095$) and at the threshold for Qwen3-32B on the train split ($p = 0.045$--$0.052$).

\begin{table}[h]
\centering
\small
\begin{tabular}{l|cc|cc}
\toprule
& \multicolumn{2}{c|}{Qwen3-8B} & \multicolumn{2}{c}{Qwen3-32B} \\
Measure & Effect & $p$ & Effect & $p$ \\
\midrule
\multicolumn{5}{l}{\textit{Aggregate}} \\
\midrule
Flip rate & $+0.005$ & ${<}\,0.001$ & $+0.003$ & $0.052$ \\
MI & $-0.035$ & ${<}\,0.001$ & $-0.021$ & $0.045$ \\
$\phi$ & $-0.009$ & ${<}\,0.001$ & $-0.006$ & $0.045$ \\
\midrule
\multicolumn{5}{l}{\textit{Per-sample}} \\
\midrule
JSD & $+0.003$ & $3.0 \times 10^{-6}$ & $+0.004$ & $4.9 \times 10^{-15}$ \\
KL & $+0.015$ & $7.2 \times 10^{-4}$ & $+0.022$ & $5.7 \times 10^{-14}$ \\
\midrule
\multicolumn{5}{l}{\textit{Regression}} \\
\midrule
$\hat{\beta}_{\text{gender}}$ & $-0.567$ & $\ll 10^{-20}$ & $-0.488$ & $\ll 10^{-20}$ \\
\bottomrule
\end{tabular}
\caption{Gender bias on bias-in-bios against the fixed sentence baseline (train split, $N{=}4{,}472$). Same format as Table~\ref{tab:bib_results_train}.}
\label{tab:bib_results_fixed_train}
\end{table}

\begin{table}[h]
\centering
\small
\begin{tabular}{l|cc|cc}
\toprule
& \multicolumn{2}{c|}{Qwen3-8B} & \multicolumn{2}{c}{Qwen3-32B} \\
Measure & Effect & $p$ & Effect & $p$ \\
\midrule
\multicolumn{5}{l}{\textit{Aggregate}} \\
\midrule
Flip rate & $+0.003$ & $0.118$ & $+0.006$ & $0.011$ \\
MI & $-0.022$ & $0.095$ & $-0.042$ & $0.010$ \\
$\phi$ & $-0.006$ & $0.095$ & $-0.013$ & $0.010$ \\
\midrule
\multicolumn{5}{l}{\textit{Per-sample}} \\
\midrule
JSD & $+0.003$ & $5.5 \times 10^{-4}$ & $+0.005$ & $6.3 \times 10^{-7}$ \\
KL & $+0.016$ & $2.4 \times 10^{-4}$ & $+0.030$ & $5.4 \times 10^{-6}$ \\
\midrule
\multicolumn{5}{l}{\textit{Regression}} \\
\midrule
$\hat{\beta}_{\text{gender}}$ & $-0.541$ & $\ll 10^{-20}$ & $-0.490$ & $\ll 10^{-20}$ \\
\bottomrule
\end{tabular}
\caption{Gender bias on bias-in-bios against the fixed sentence baseline (test split, $N{=}1{,}724$). Same format as Table~\ref{tab:bib_results_train}.}
\label{tab:bib_results_fixed_test}
\end{table}

\section{Q-Pain Replication: Full Setup and Results}
\label{app:qpain_replication}

We replicate the original Q-Pain study's full experimental design \citep{loge_q-pain_2021}, adding our adjusted paraphrase and fixed sentence baselines. The Q-Pain dataset contains 50 clinical vignettes across five pain contexts (acute cancer, acute non-cancer, chronic cancer, chronic non-cancer, and post-operative), each with demographic placeholders for race, gender, and patient name. The original Q-Pain study found that GPT-2 Large exhibited significant racial disparities, particularly disadvantaging Asian men relative to other subgroups.

\paragraph{Setup.}
Following the original Q-Pain code, we evaluate GPT-2 Large on all 8 demographic subgroups (4 races $\times$ 2 genders) using the paper's few-shot prompting format with 2 closed examples (Patient~A and Patient~B). The original code comments out the third closed prompt (Patient~C, a ``Yes, Low dose'' example), although the paper reports that there are 3 closed examples; we follow the code. 
We extract the probability of treatment denial ($p(\text{No})$) from the vocabulary distribution (after softmax) at the first token position after ``Answer:''. For each of the $\binom{8}{2} = 28$ pairwise subgroup comparisons, we conduct a paired two-tailed $t$-test (paired by vignette, $N{=}50$) on $p(\text{No})$ differences, with Bonferroni correction ($\alpha = 0.05/28 \approx 0.002$). This matches the original study's statistical methodology.

For each comparison, we randomly designate one subgroup as ``original'' and generate an adjusted paraphrase of that subgroup's vignette targeting the token edit distance to the other subgroup's vignette (measured with the GPT-2 tokenizer). For the fixed sentence baseline, we prepend ``This patient record has been reviewed.'' to the original subgroup's vignette. Both baselines are evaluated with the same paired $t$-test framework.

\paragraph{Replication fidelity.}
Our replication finds 5 of 28 comparisons significant after Bonferroni correction, compared to 7 in the original study. 
All 7 of the original study's significant comparisons are significant at $p < 0.05$ in our replication \emph{without correcting for multiple hypothesis testing}; 3 of the 7 also pass the Bonferroni threshold. 
Both analyses identify Asian men as the most disadvantaged subgroup. 

\paragraph{Results.}

\begin{table}[h]
\centering
\begin{tabular}{l|ccc}
\toprule
Test family & Comparisons & Sig.\ ($p < 0.05$) & Sig.\ (Bonferroni) \\
\midrule
Demographic & 28 & 13 & 5 \\
adjusted paraphrase & 28 & 28 & 18 \\
Fixed sentence & 8 & 8 & 8 \\
\bottomrule
\end{tabular}
\caption{Summary of paired $t$-tests on $p(\text{No})$ for the Q-Pain replication. Demographic and paraphrase tests each have 28 unique comparisons (Bonferroni $\alpha = 0.05/28$). The fixed sentence test depends only on the ``original'' subgroup---it compares a subgroup's demographic $p(\text{No})$ against the same subgroup's fixed-sentence $p(\text{No})$, which does not depend on the comparison partner---yielding 8 unique tests (Bonferroni $\alpha = 0.05/8$).}
\label{tab:qpain_replication_summary}
\end{table}

Table~\ref{tab:qpain_replication_summary} presents the summary. Three of the 5 Bonferroni-significant demographic comparisons involve Asian men being denied treatment at higher rates: Hispanic Man vs.\ Asian Man ($\Delta = -0.029$, $p < 10^{-6}$), Black Man vs.\ Asian Man ($\Delta = -0.025$, $p < 10^{-4}$), and Asian Woman vs.\ Asian Man ($\Delta = -0.022$, $p < 0.001$); the remaining two are White Man vs.\ Hispanic Man ($\Delta = +0.018$) and Hispanic Man vs.\ White Woman ($\Delta = -0.021$). This is consistent with the original study's finding that GPT-2 ``mostly singled out Asian men.''

Crucially, 4 of the 5 Bonferroni-significant demographic comparisons have a corresponding Bonferroni-significant adjusted paraphrase effect (the exception is Hispanic Man vs.\ Asian Man, whose paraphrase effect is significant at $p < 0.05$ but not after Bonferroni correction). All 5 have Bonferroni-significant fixed sentence counterparts. This means that for nearly every demographic comparison where we detect a significant effect, we also detect a significant effect from simply paraphrasing or prepending a sentence---casting doubt on whether the demographic effect is attributable to race specifically.

More broadly, both baselines show greater overall sensitivity than the demographic swaps. All 28 adjusted paraphrase comparisons are significant at $p < 0.05$ (18 at Bonferroni), and all 8 unique fixed sentence tests are Bonferroni-significant. This indicates that GPT-2 Large's output probabilities are sensitive to any text perturbation, not specifically to demographic content.

\section{Q-Pain Regression Experiment: Full Setup and Results}
\label{app:qpain_regression}

The Q-Pain dataset contains 50 clinical vignettes across five pain contexts (acute cancer, acute non-cancer, chronic cancer, chronic non-cancer, and post-operative), each with demographic placeholders for race, gender, and patient name. The original Q-Pain study found that GPT-2 Large exhibited significant racial disparities, particularly disadvantaging Asian men relative to other subgroups. However, a subsequent study \citep{poulain_bias_2024} testing eight modern LLMs found no statistically significant global bias on the same dataset.

We test two race comparisons---Black vs.\ White and Asian vs.\ White---across three models: GPT-2 Large (774M, the model from the original Q-Pain study), Qwen3-8B, and Qwen3-32B. For each comparison, the directional hypothesis is $H_1{:}\ \beta_{\text{race}} < 0$ (non-White patients denied treatment more). White serves as the baseline in all comparisons.

Following the original Q-Pain vignette construction protocol, we fill each template with race, gender (fixed to male), pronouns, and racially-associated patient names drawn from the Harvard Dataverse demographic data used in the original study. For Qwen3 models, each vignette is presented with its clinical question in a binary-choice format (``Respond with only Yes or No'') and we extract the log-odds $\text{logit}_{\text{Yes}} - \text{logit}_{\text{No}}$ from next-token logits. For GPT-2 Large, we use the original code's few-shot prompting format (two closed examples followed by the open prompt) and extract log-odds from the first token position after ``Answer:''. Both baselines (adjusted paraphrase and fixed sentence) are included; paraphrases are adjusted separately per model tokenizer to match the token edit distance of each race swap.

\begin{table}[h]
\centering
\begin{tabular}{ll|cc|cc}
\toprule
& & \multicolumn{2}{c|}{Black vs.\ White} & \multicolumn{2}{c}{Asian vs.\ White} \\
Model & Baseline & $\hat{\beta}_{\text{race}}$ & $p$ & $\hat{\beta}_{\text{race}}$ & $p$ \\
\midrule
\multirow{2}{*}{GPT-2 Large}
    & Paraphrase & $+0.117$ & $0.999$ & $-0.004$ & $0.427$ \\
    & Fixed sent. & $+0.034$ & $0.819$ & $\mathbf{-0.087}$ & $\mathbf{< 0.001}$ \\
\multirow{2}{*}{Qwen3-8B}
    & Paraphrase & $+0.225$ & $0.735$ & $-0.110$ & $0.365$ \\
    & Fixed sent. & $+0.535$ & $0.944$ & $+0.315$ & $0.778$ \\
\multirow{2}{*}{Qwen3-32B}
    & Paraphrase & $+0.318$ & $0.972$ & $+0.080$ & $0.688$ \\
    & Fixed sent. & $+0.250$ & $0.921$ & $-0.007$ & $0.482$ \\
\bottomrule
\end{tabular}
\caption{Regression results for race bias on Q-Pain ($N{=}50$). $\hat{\beta}_{\text{race}}$ estimates the directional shift in treatment log-odds when patient race is changed to the non-White group. One-sided $p$-values test $H_1{:}\ \beta_{\text{race}} < 0$. \textbf{Bold} indicates $p < 0.05$. The level model yields identical $\hat{\beta}_{\text{race}}$ estimates and is omitted.}
\label{tab:qpain_results}
\end{table}

Table~\ref{tab:qpain_results} presents the results across 3 models $\times$ 2 comparisons $\times$ 2 baselines (12 conditions). Of these, only one reaches significance: GPT-2 Large on the Asian-vs.-White comparison against the fixed sentence baseline ($\hat{\beta}_{\text{race}} = -0.087$, $p < 0.001$). However, the same comparison against the adjusted paraphrase baseline---which matches the perturbation magnitude of the race swap---is not significant ($p = 0.427$). All Black-vs.-White conditions and all Qwen3 conditions are non-significant.

The contrast with Bias-in-Bios is informative. The same regression framework that detects highly significant gender-occupation bias ($\hat{\beta}_{\text{gender}} \approx -0.5$, $p < 10^{-100}$) against both baselines produces largely null results for race bias in pain management. The one significant GPT-2 result does not survive the adjusted paraphrase baseline, and neither modern model shows any race effect against either baseline.

\clearpage
\section{Adjusted Paraphrase Baselines: Accuracy and Examples}
\label{app:paraphrase_calibration}

\paragraph{Procedure.} For the MedPerturb experiments, paraphrases are generated independently for each (vignette, perturbation) pair, each adjusted to match that specific perturbation's token edit distance for that specific vignette. The target percentage is the Levenshtein distance on the tokenized sequences divided by the original token count. The paraphrasing model ({\tt GPT-5.2}) receives iterative feedback on the measured token change in a multi-turn loop of up to 50 attempts; if no attempt falls within $\pm 0.5$ percentage points (pp) of the target, we select the closest attempt that does not exceed the target. In a small number of instances where every attempt overshot the target, we took a second pass in which we instructed the model to perform synonym replacements only. 
Counts below are over the 400 baselines used in the MedPerturb analysis (100 vignettes $\times$ 4 perturbations).

\paragraph{Accuracy.} Table~\ref{tab:paraphrase_accuracy} reports how closely the generated baselines match their targets. Matching is near-exact for small targets and degrades for large ones: the uncertain and colorful perturbations rewrite the vignette extensively (median targets of 57\% and 53\% of tokens, with maxima of 331\% and 571\%), which makes an edit-distance-matched paraphrase difficult to construct at all.

\begin{table}[t]
\centering
\small
\begin{tabular}{lccc}
\toprule
Target range & $N$ & Within $\pm 0.5$pp & Misses (all undershoots) \\
\midrule
$\leq 5\%$ (gender swap/remove) & 157 & 153 (97.5\%) & 4 (mean dev.\ 1.6pp) \\
5--20\% & 47 & 47 (100\%) & 0 \\
$> 20\%$ (uncertain, colorful) & 196 & 43 (21.9\%) & 153 (mean dev.\ 27.4pp) \\
\bottomrule
\end{tabular}
\caption{Accuracy of the adjusted paraphrase baseline over the 400 baselines used in the MedPerturb experiments. ``Target range'' is the token change percentage of the targeted perturbation that the paraphrase is adjusted to match; the $>20\%$ row comprises 100 uncertain, 94 colorful, and 2 gender-remove cases. Every miss outside the tolerance is an undershoot: no paraphrase exceeds its target by more than the tolerance.}
\label{tab:paraphrase_accuracy}
\end{table}

\paragraph{Direction of the residual mismatch.} Every miss is an undershoot; no paraphrase exceeds its target by more than the tolerance. 
This is because our selection rule prefers the closest attempt that does not exceed the target. The result is that paraphrases used tend to have fewer edited tokens than the matched targeted perturbation; the baseline is thus less aggressive than the targeted perturbation it is compared against, making it relatively ``permissive''. 
This strengthens the reported null results; failure to reject is not an artifact of the baseline editing substantially (relative to the targeted edit). 
For the gender perturbation results (which were null), 196 of 200 baselines (98\%) are within tolerance. 

For the one perturbation that does reach significance against this baseline---colorful on the 8B model (Table~\ref{tab:calibrated_results})---the same observation has the opposite implication: it biases the perturbation$-$baseline difference upward. 
But the mismatch in edited tokens is small for most vignettes: Half of the colorful baselines achieve at least 97\% of their targets (a median shortfall of 1.2pp), and the bucket's large mean deviation is driven by a small tail of extreme targets, so we do not expect the mismatch to account for the effect. 
This is the same token-mismatch confound we discuss for the fixed sentence baseline (Appendix~\ref{app:full_medperturb}) in weaker form; the colorful result should be interpreted with this in mind.

\paragraph{Examples.} Table~\ref{tab:paraphrase_examples} shows one successful match and one undershoot, both for the same vignette (58 tokens). 
For the gender swap the target is a single token edit, which the paraphraser matches exactly on its first attempt. 
For the colorful perturbation the target is 222 token operations---nearly four times the length of the original, and among the largest targets in the set---and the closest non-exceeding attempt after 50 retries reaches 142 (off by 138pp).

\begin{table}[t]
\centering
\footnotesize
\begin{tabular}{@{}p{0.13\textwidth}p{0.81\textwidth}@{}}
\toprule
\multicolumn{2}{@{}l}{\textbf{(a) Successful match.} Gender swap.} \\
\midrule
Original & My arms sometimes hurt when I sneeze? 28F, ex smoker, no drinking, 5'3'', 200lbs. Its not always, just sometimes I get a \emph{somewhat} intense ache down either or both arms right after I sneeze. Should I be worried?! \\
\addlinespace
Target & 1.7\% (1 of 58 tokens) \\
Achieved & 1.7\% (1 token; 0 retries) \\
\addlinespace
Paraphrase & My arms sometimes hurt when I sneeze? 28F, ex smoker, no drinking, 5'3'', 200lbs. Its not always, just sometimes I get a \emph{fairly} intense ache down either or both arms right after I sneeze. Should I be worried?! \\
\midrule
\multicolumn{2}{@{}l}{\textbf{(b) Undershoot.} Colorful, same vignette.} \\
\midrule
Target & 382.8\% (222 token operations) \\
Achieved & 244.8\% (142 operations; 50 retries) \\
\addlinespace
Paraphrase & I'm looking for guidance about a symptom that shows up only in connection with sneezing. I'm 28F, an ex smoker, I do not drink, and my height/weight are 5'3'' and 200lbs. Most of the time when I sneeze nothing unusual happens, but once in a while there's an immediate follow-up sensation: a somewhat intense ache that seems to run downward along one arm. The side isn't predictable---sometimes it's the left, sometimes the right---and on certain occasions it feels like it's in both arms simultaneously. The discomfort starts right after the sneeze rather than before it, and it's more of an ache than a sharp stab. With this happening only intermittently yet occurring right after I sneeze and involving either or both arms, should I be worried or consider this urgent? \\
\bottomrule
\end{tabular}
\caption{Adjusted paraphrase baselines for one vignette under two perturbations. In (a), the targeted perturbation changes one token, and the adjusted paraphrase changes one token (italicised). In (b), the targeted perturbation is an extensive rewrite; the paraphraser cannot reach the target, and we retain the closest attempt that does not exceed it. Percentages are token edit distance over the original's 58 tokens.}
\label{tab:paraphrase_examples}
\end{table}

\clearpage
\section{Self-Consistency as a Reference Distribution}
\label{app:self_consistency}

The paraphrase baseline asks whether a targeted perturbation changes the model's output distribution more than a benign rewording of the same magnitude does. A weaker reference precedes it: even with the prompt held fixed, repeated sampling produces variation in the model's answers, and a targeted effect should at minimum exceed this sampling-noise floor. We therefore also test each MedPerturb perturbation against a self-consistency reference, which holds the prompt fixed and varies only the sampling noise.

\paragraph{Setup.} For each of the 100 MedPerturb cases, each of the three tasks, and each condition (the original prompt and the four targeted perturbations), we draw $N = 10$ i.i.d.\ samples at temperature $0.7$ and record the binary answer of each. This yields an empirical answer distribution $\hat{P}_i$ per (case, condition, task), in place of the three-sample majority vote used for the thresholded metrics in Section~\ref{section:revisit-medperturb} (see \emph{Interpreting the comparison} below). For each case we compute the Bernoulli JSD (base 2) between the original and perturbed empirical distributions, and take the mean across the 100 cases as the test statistic. Every test uses all 100 cases. For many cases, however, the model returns the same answer in all $2N = 20$ pooled samples---the $N$ samples drawn for the original prompt together with the $N$ drawn for its perturbed counterpart; such a unanimous case contributes exactly zero to the test: its two empirical distributions are identical, so its observed JSD is zero, and since every permutation merely reshuffles 20 identical answers, its JSD is zero under the null as well. Only cases with at least one dissenting answer among the 20 pooled samples carry information; the number of these informative cases per cell ranges from 7 to 80 and is reported in Table~\ref{tab:self_consistency}.

\paragraph{Statistical testing.} The null hypothesis is that the perturbed samples are exchangeable with the original samples---that is, that the observed divergence is no larger than what repeated sampling of the unchanged prompt would produce. Unlike the tests in Section~\ref{sec:testing}, which contrast a targeted perturbation against a baseline perturbation, the self-consistency null is an exchangeability null, which admits an exact permutation test. For each case we pool its $2N$ samples, reassign them at random to the two conditions, and recompute the mean JSD across cases. The $p$-value is $(1 + \#\{\text{null} \geq \text{observed}\})/(1 + B)$ with $B = 10{,}000$ permutations; the $+1$ in numerator and denominator makes the test exact under the exchangeability null. As in the main experiment, we apply a Bonferroni correction within each group of 12 tests (4 perturbations $\times$ 3 tasks, $\alpha = 0.05/12 \approx 0.004$), separately per model.

\begin{table}[t]
\centering
\small
\begin{tabular}{l|cc|cc}
\toprule
& \multicolumn{2}{c|}{Significant tasks} & \multicolumn{2}{c}{Informative cases} \\
Perturbation & Llama 3.1 8B & Llama 3.1 70B & 8B & 70B \\
\midrule
Gender-swap   & 0 / 3 & 2 / 3 & 12--76 & 7--21 \\
Gender-remove & 0 / 3 & 0 / 3 & 10--78 & 7--19 \\
Uncertain     & 3 / 3 & 3 / 3 & 11--80 & 7--27 \\
Colorful      & 3 / 3 & 3 / 3 & 10--79 & 9--22 \\
\bottomrule
\end{tabular}
\caption{Perturbation effects tested against the self-consistency reference ($N = 100$ cases; 10 samples per prompt at temperature $0.7$). ``Significant tasks'' reports the number of the three triage tasks (MANAGE, VISIT, RESOURCE) on which the perturbation produces significantly greater divergence than repeated sampling of the unchanged prompt (within-case permutation test, $B = 10{,}000$; $p < 0.05/12$, Bonferroni-corrected). ``Informative cases'' gives the range, across the three tasks, of the number of cases whose pooled samples are not unanimous; unanimous cases contribute zero to both the observed and the null statistic.}
\label{tab:self_consistency}
\end{table}

\paragraph{Results.} Table~\ref{tab:self_consistency} reports the results. The uncertain and colorful perturbations exceed sampling noise on every task for both models (3/3 each). Gender-remove does so on no task for either model. Gender-swap reaches significance on two of the three tasks on the 70B model (MANAGE and RESOURCE, but not VISIT) and on none on the 8B model. Model size may be one reason for this difference. Each test measures divergence in excess of the model's own sampling noise, and the two models' noise floors differ: the 70B's repeated samples are less variable in every cell of this experiment. On MANAGE, for instance, the two models' observed divergences are nearly identical (0.0201 vs.\ 0.0192), but the 70B's null mean is less than half the 8B's, so the same absolute movement is significant for the 70B and not for the 8B. The within-model original-versus-perturbed contrast, which is what each test evaluates, is unaffected by this difference.

\paragraph{Comparison with the paraphrase baseline.} Fourteen of the 24 (perturbation, task, model) cells reject the self-consistency null. We call a cell paraphrase-significant if any of the five metrics is significant against the adjusted paraphrase baseline; this is the permissive reading, and it works against the comparison below---under the stricter rule requiring all five metrics, no cell is paraphrase-significant at all. Cross-referencing the 14 cells with the adjusted paraphrase baseline results (Table~\ref{tab:calibrated_results}), where only 5 of 120 tests reach significance---all for colorful on the 8B model---reveals three patterns.

\emph{(i) Both references reject.} Colorful on the 8B model, for MANAGE and VISIT. These effects are robust against both sampling noise and paraphrase variation.

\emph{(ii) Only the self-consistency reference rejects.} The remaining 12 self-consistency-significant cells: uncertain (all tasks, both models), colorful on the 70B model (all tasks), colorful on the 8B model for RESOURCE, and gender-swap on the 70B model (MANAGE and RESOURCE). These effects exceed sampling noise but are indistinguishable from the surface-form sensitivity induced by paraphrasing.

\emph{(iii) Neither rejects.} Gender-swap on the 8B model, gender-remove on both models, and gender-swap on the 70B model for VISIT.

Pattern (ii) dominates: 12 of the 14 cells that reject the self-consistency null are filtered out by the paraphrase baseline. This suggests that self-consistency alone would substantially overcount ``real'' perturbation effects. The paraphrase baseline provides the additional discriminating power, filtering out effects that exceed sampling noise but can be explained by general prompt sensitivity rather than by the targeted factor. The two controls are complementary rather than interchangeable: the self-consistency reference controls for output stochasticity, while the paraphrase baseline controls for input confounding.

\paragraph{Interpreting the comparison.} The two tests are not built on the same statistic, and cannot be. The self-consistency test uses an empirical distribution over 10 sampled generations, whereas the JSD and KL entries of Table~\ref{tab:calibrated_results} are computed from the next-token probabilities of a single forward pass---which, for a fixed prompt, are deterministic and therefore have no sampling noise to test against---and the flip rate, MI, and $\phi$ entries from a three-seed majority vote. The self-consistency reference is thus necessarily restricted to the sampled answers. The two tests also differ in structure: the self-consistency test asks whether the perturbed and original samples are distinguishable at all, whereas the paraphrase test pairs each case's targeted divergence with its baseline divergence and asks whether the difference is positive on average (Section~\ref{sec:testing}), a more demanding comparison. The filtering reported above---the 12 of 14 self-consistency-significant cells that do not reject under the paraphrase test---therefore reflects the change in reference distribution together with these differences in statistic and in test structure, and we do not attempt to determine how much each contributes.

\clearpage
\section{Bootstrap Test Specification}
\label{app:bootstrap}

This appendix specifies the nonparametric bootstrap summarized in Section~\ref{sec:testing}, which we use to test the aggregate metrics (flip rate, MI, $\phi$). These metrics yield a single value per dataset, so there are no per-sample differences to test directly; the bootstrap supplies the sampling distribution of the targeted-minus-baseline difference instead.

\paragraph{Resampling unit.} The resampling unit is the individual case, i.e., an original prompt $p_i$ together with its targeted counterpart $p'_i$ and its baseline counterpart $p''_i$. Each bootstrap iteration draws $N$ cases with replacement from the $N$ original cases, where $N$ is the dataset size.

\paragraph{Paired resampling.} The targeted and baseline conditions are resampled \emph{jointly}: each iteration draws $N$ cases with replacement, and the same drawn cases, with the same multiplicities, form both resampled datasets---$\mathcal{D}^*_{\text{targeted}}$ collects the pairs $(p_j, p'_j)$ and $\mathcal{D}^*_{\text{benign}}$ the pairs $(p_j, p''_j)$. The bootstrap replicate of the difference is
\begin{equation}
\hat{\delta}^* = \Psi(\mathcal{D}^*_{\text{targeted}}, f) - \Psi(\mathcal{D}^*_{\text{benign}}, f),
\end{equation}
which instantiates the definition in Section~\ref{sec:testing}: $\hat{\Psi}^*_{\text{targeted}} = \Psi(\mathcal{D}^*_{\text{targeted}}, f)$ and $\hat{\Psi}^*_{\text{baseline}} = \Psi(\mathcal{D}^*_{\text{benign}}, f)$. Each resampled case thus contributes its targeted and its baseline observation together; resampling the two conditions on independent draws would discard this pairing and inflate the variance of $\hat{\delta}^*$.

\paragraph{Bootstrap distribution.} We draw $B = 1{,}000$ replicates of $\hat{\delta}^*$, giving the bootstrap distribution of the difference between the targeted and the baseline value of $\Psi$. The preliminary MedQA experiment of Table~\ref{tab:preliminary_results} uses $B = 10{,}000$.

\paragraph{Confidence intervals.} The same replicates yield a 95\% confidence interval for the targeted-minus-baseline difference by the percentile method: the endpoints are the 2.5th and the 97.5th percentiles of the $B$ replicates. This interval is two-sided and is reported as a descriptive summary of the bootstrap distribution; significance is determined by the one-sided $p$-value below, not by whether the interval covers zero. The interval corresponding exactly to the one-sided test at level $\alpha$ is its one-sided analogue, bounded by the $\alpha$ (or $1-\alpha$) quantile of the replicates.

\paragraph{$p$-values.} The $p$-value is one-sided and matches the directional alternative of the metric, i.e., it is the proportion of replicates falling on the side of zero that is inconsistent with $H_1$: $\Pr(\hat{\delta}^* \le 0)$ for flip rate, and $\Pr(\hat{\delta}^* \ge 0)$ for MI and $\phi$. The directions differ because the metrics point in opposite ways: a larger flip rate indicates a larger perturbation effect, whereas larger MI or $\phi$ indicates \emph{more} agreement between the original and perturbed answers, and hence a \emph{smaller} effect. A one-sided formulation matches these directional hypotheses; the two-sided alternative would spend power on differences in the direction we have no hypothesis for.

\clearpage
\section{DiscrimEval: Full Setup and Per-Contrast Results}
\label{app:discrimeval}

DiscrimEval \citep{tamkin_evaluating_2023} contains templated decision scenarios in which a demographic profile (age, race, gender) is filled into a prompt asking for a binary yes/no recommendation. 
We use the 70 ``Explicit'' scenarios, in which demographic attributes are stated directly in the text. Following the benchmark's own construction, we fix the white, 60-year-old, male profile as the reference and form 14 contrasts against it: 2 gender  ({\tt gender\_female}, {\tt gender\_non\_binary}), 4 race  ({\tt race\_Black}, {\tt race\_Asian}, {\tt race\_Hispanic}, {\tt race\_Native\_American}), and 8 age contrasts ({\tt age\_20} through {\tt age\_100}, excluding the reference age).

For each of the $70 \times 14 = 980$ (scenario, contrast) pairs we generate an adjusted paraphrase of the reference scenario, matching the token edit distance of that specific contrast for that specific scenario, following the procedure outlined in  Section~\ref{sec:baselines}. 
We then obtain three log-odds values per pair---reference, contrast, and paraphrase---as $\text{logit}_{\text{Yes}} - \text{logit}_{\text{No}}$ at the first generated token position, and compute all five metrics using the induced Bernoulli distributions. 
We test JSD and KL with one-sided paired $t$-tests over the 70 scenarios. We test flip rate, MI, and $\phi$ with the scenario-level bootstrap described in  Section~\ref{sec:testing} ($B = 1{,}000$). Significance is assessed at $\alpha = 0.05/14 \approx 0.0036$.

\begin{table}[t]
\centering
\footnotesize
\setlength{\tabcolsep}{6pt}
\begin{tabular}{l|cc|cc}
\toprule
& \multicolumn{2}{c|}{Llama 3.1 8B} & \multicolumn{2}{c}{Llama 3.1 70B} \\
Axis & Raw effect & Surviving & Raw effect & Surviving \\
\midrule
Gender & $+0.20$ & 0 / 2 & $+0.51$ & 0 / 2 \\
Race   & $+0.20$ & 0 / 4 & $+0.84$ & 0 / 4 \\
Age    & $-0.04$ & 0 / 8 & $-0.24$ & 0 / 8 \\
\midrule
Total  & ---     & 0 / 14 & ---    & 0 / 14 \\
\bottomrule
\end{tabular}
\caption{{\tt DiscrimEval} results against the adjusted paraphrase baseline (70 scenarios $\times$ 14 contrasts). Raw effect is the mean over the 70 scenarios of the difference in $\text{logit}_{\text{Yes}} - \text{logit}_{\text{No}}$ between the contrast and the reference profile---the log-odds parameter of each scenario's Bernoulli distribution, on which our five metrics are computed---averaged over the contrasts on each axis. It is reported in the same units as the discrimination score of \citet{tamkin_evaluating_2023} to permit direct comparison with the original audit; it is not itself one of our test statistics. ``Surviving'' counts contrasts with at least one of the five metric cells reaching $p < 0.05/14 \approx 0.0036$ under our paraphrase-controlled test. Per-contrast results are in Table~\ref{tab:discrimeval_full}.}
\label{tab:discrimeval_results}
\end{table}

Raw demographic effects are substantial on both models, and larger on the 70B: {\tt race\_Black} shows a raw effect of $+1.00$ log-odds relative to the reference, {\tt gender\_non\_binary} $+0.70$, and there is an overall negative age gradient from {\tt age\_20} ($+0.06$) to {\tt age\_100} ($-0.72$). These patterns qualitatively match those \citet{tamkin_evaluating_2023} report on Claude 2.0, and taken alone they might be read as supporting the same discrimination conclusion. Against the adjusted paraphrase baseline, however---a control not used in the original analysis---no contrast $\times$ metric cell reaches Bonferroni significance on either model: 0 of 14 contrasts survive (Table~\ref{tab:discrimeval_results}). On Llama 3.1, the demographic signal is statistically indistinguishable from the model's response to a same-magnitude paraphrase; this is a re-analysis of the benchmark on a different model, not a contradiction of the original Claude 2.0 result.

Each contrast-level test uses $n = 70$ paired observations, so these nulls assert statistical indistinguishability from the baseline. They are not artifacts of insensitive metrics, however: the null holds for JSD and KL, the metrics our power analysis (Appendix~\ref{app:power_curves}) finds most sensitive, on raw shifts that reach roughly one log-odds unit on the 70B.

Each contrast is a categorical comparison against a single reference profile rather than a signed perturbation of a fixed attribute, so we report the non-directional tests here; the regression specification of Section~\ref{sec:testing} does not apply as directly as it does on {\tt bias-in-bios}.

Table~\ref{tab:discrimeval_full} reports, for each contrast, the raw effect and the smallest $p$-value across the five metrics. The raw effect is the mean difference in $\text{logit}_{\text{Yes}} - \text{logit}_{\text{No}}$ between the contrast and the reference over the 70 scenarios; it is reported in the same units as the discrimination score of \citet{tamkin_evaluating_2023} to allow direct comparison with the original audit (this is not one of our test statistics we proposed). No cell is significant on either model: the smallest $p$-value observed over all five metrics is $0.010$ on the 8B ({\tt gender\_female}) and $0.007$ on the 70B ({\tt age\_20}, JSD), each a factor of $2$--$3$ above the corrected threshold. The age gradient is negative overall but not monotonic on either model.

\begin{table}[h]
\centering
\footnotesize
\setlength{\tabcolsep}{6pt}
\begin{tabular}{ll|cc|cc}
\toprule
& & \multicolumn{2}{c|}{Llama 3.1 8B} & \multicolumn{2}{c}{Llama 3.1 70B} \\
Axis & Contrast & Raw effect & $\min p$ & Raw effect & $\min p$ \\
\midrule
\multirow{2}{*}{Gender}
    & {\tt gender\_female}        & $+0.21$ & $.010$ & $+0.33$ & $.048$ \\
    & {\tt gender\_non\_binary}   & $+0.20$ & $.041$ & $+0.70$ & $.161$ \\
\midrule
\multirow{4}{*}{Race}
    & {\tt race\_Black}           & $+0.25$ & $.033$ & $+1.00$ & $.032$ \\
    & {\tt race\_Asian}           & $+0.15$ & $.075$ & $+0.69$ & $.155$ \\
    & {\tt race\_Hispanic}        & $+0.23$ & $.050$ & $+0.77$ & $.161$ \\
    & {\tt race\_Native\_American}& $+0.19$ & $.077$ & $+0.89$ & $.089$ \\
\midrule
\multirow{8}{*}{Age}
    & {\tt age\_20}               & $-0.09$ & $.089$ & $+0.06$ & $.007$ \\
    & {\tt age\_30}               & $-0.04$ & $.338$ & $-0.05$ & $.198$ \\
    & {\tt age\_40}               & $-0.05$ & $.363$ & $+0.10$ & $.039$ \\
    & {\tt age\_50}               & $+0.00$ & $.131$ & $+0.02$ & $.128$ \\
    & {\tt age\_70}               & $+0.04$ & $.151$ & $-0.20$ & $.464$ \\
    & {\tt age\_80}               & $+0.02$ & $.112$ & $-0.54$ & $.047$ \\
    & {\tt age\_90}               & $-0.06$ & $.351$ & $-0.58$ & $.042$ \\
    & {\tt age\_100}              & $-0.16$ & $.144$ & $-0.72$ & $.044$ \\
\bottomrule
\end{tabular}
\caption{Per-contrast DiscrimEval results against the adjusted paraphrase baseline ($n = 70$ scenarios per contrast). Raw effect is the mean difference in $\text{logit}_{\text{Yes}} - \text{logit}_{\text{No}}$ between the contrast and the reference profile. ``$\min p$'' is the smallest one-sided $p$-value across the five metrics. No contrast reaches the Bonferroni-corrected threshold $p < 0.05/14 \approx 0.0036$ on either model; all 70 contrast $\times$ metric cells are non-significant for both models.}
\label{tab:discrimeval_full}
\end{table}

\end{document}